\documentclass[10pt,twocolumn,letterpaper]{article}

\usepackage[pagenumbers]{wacv} 

\usepackage{graphicx}
\usepackage{amsmath}
\usepackage{amssymb}
\usepackage{booktabs}

\usepackage{multirow}
\usepackage{caption}
\usepackage{subcaption}
\usepackage{makecell}
\usepackage{color, colortbl}
\definecolor{Gray}{gray}{0.9}

\usepackage[pagebackref,breaklinks,colorlinks]{hyperref}

\usepackage[capitalize]{cleveref}
\crefname{section}{Sec.}{Secs.}
\Crefname{section}{Section}{Sections}
\Crefname{table}{Table}{Tables}
\crefname{table}{Tab.}{Tabs.}

\def\wacvPaperID{1789} 
\def\confName{WACV}
\def\confYear{2024}

\begin{document}

\title{Limited Data, Unlimited Potential: \\ A Study on ViTs Augmented by Masked Autoencoders}

\author{Srijan Das$^{1}$, Tanmay Jain$^{2}$, Dominick Reilly$^{1}$, Pranav Balaji$^{3}$, Soumyajit Karmakar$^{4}$,\\ Shyam Marjit$^{4}$, Xiang Li$^{5}$, Abhijit Das$^{3}$, and Michael S.  Ryoo$^{5}$\\
$^{1}$UNC Charlotte, 
$^{2}$ Delhi Technological University, 
$^{3}$ BITS Pilani Hyderabad\\
$^{4}$ Indian Institute of Information Technology Guwahati, 
$^{5}$ Stony Brook University \\
{\tt\small sdas24@charlotte.edu}\\
}

\maketitle

\begin{abstract}
   Vision Transformers (ViTs) have become ubiquitous in computer vision. Despite their success, ViTs lack inductive biases, which can make it difficult to train them with limited data. To address this challenge, prior studies suggest training ViTs with self-supervised learning (SSL) and fine-tuning sequentially. 
   However, we observe that jointly optimizing ViTs for the primary task and a Self-Supervised Auxiliary Task (SSAT) is surprisingly beneficial when the amount of training data is limited.
   We explore the appropriate SSL tasks that can be optimized alongside the primary task, the training schemes for these tasks, and the data scale at which they can be most effective. Our findings reveal that SSAT is a powerful technique that enables ViTs to leverage the unique characteristics of both the self-supervised and primary tasks, 
   achieving better performance than typical ViTs pre-training with SSL and fine-tuning sequentially. Our experiments, conducted on 10 datasets, demonstrate that SSAT significantly improves ViT performance while reducing carbon footprint. We also confirm the effectiveness of SSAT in the video domain for deepfake detection, showcasing its generalizability. Our code is available at \url{https://github.com/dominickrei/Limited-data-vits}.  
\end{abstract}
\vspace{-0.1in}
\section{Introduction}
\label{sec:intro}
Vision Transformers (ViTs) have become a common sight in computer vision owing to their success across various visual tasks, and are now considered a viable alternative to Convolutional Neural Networks (CNNs). Despite this, ViTs are structurally deficient in inductive bias compared to CNNs, which necessitates training them with large-scale datasets to achieve acceptable visual representation, as noted by Dosovitskiy et al.~\cite{dosovitskiy2020vit}. As a result, when dealing with small-scale datasets, it is essential to utilize a ViT pre-trained on a large-scale dataset such as ImageNet~\cite{imagenet_cvpr09} or JFT-300M~\cite{jft_300m}. However, in domains such as medical datasets, pre-training ViTs on ImageNet or JFT-300M may not result in an optimal model for fine-tuning on those datasets due to a significant domain gap. Thus, the aim of this research is to address the following question: \textit{how can ViTs be trained effectively in domains with limited data}?

\begin{figure}
\centering
\includegraphics[width=1\linewidth]{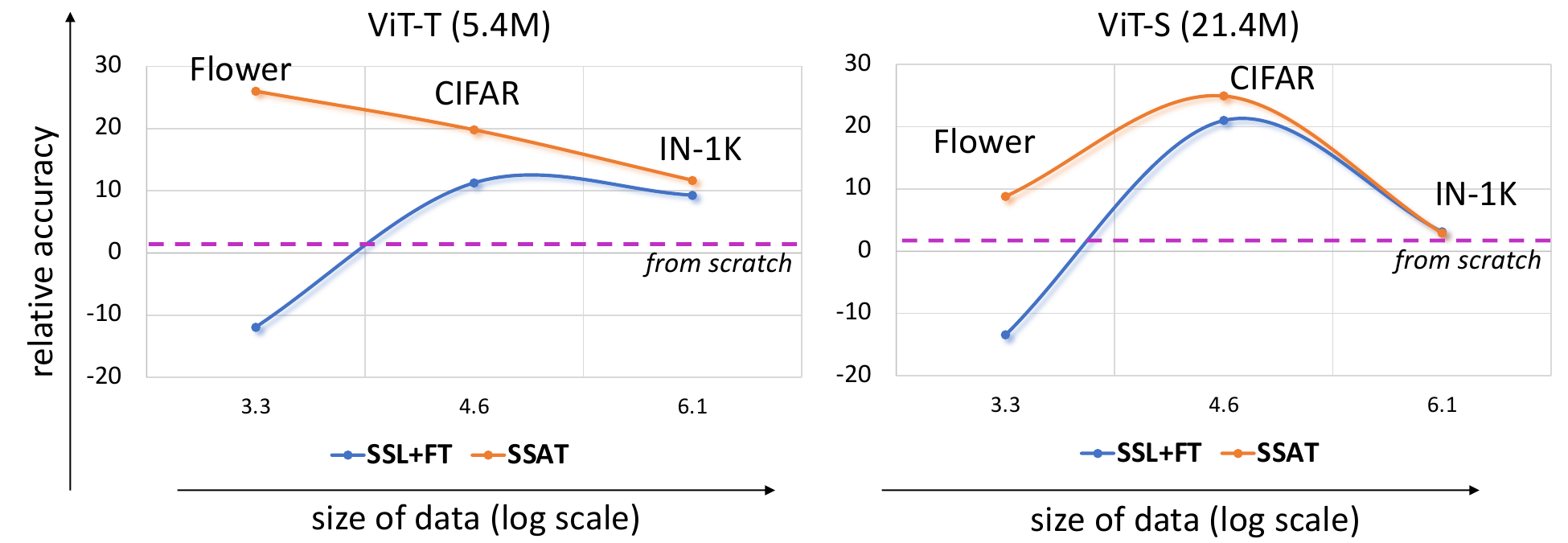}
\caption{
Relative classification accuracy on three datasets with different sizes: (i) Oxford Flower~\cite{flowers} (2K samples), (ii) CIFAR~\cite{cifar} (50K samples), and (iii) ImageNet-1K~\cite{imagenet_cvpr09} (IN-1K, 1.2M samples). SSAT consistently outperforms others on all three datasets with two backbones. On the other hand, given the same SSL method, SSL+FT achieves a compromised performance than SSAT, especially on the tiny Oxford Flower dataset (even worse than training from scratch). 
} \vspace{-0.1in}
\label{fig:intro}
\vspace{-5pt}
\end{figure}

Following the introduction of ViTs, second-generation vision transformers have emerged with two different approaches. The first approach is to use a hierarchical structure to introduce inductive bias in ViTs~\cite{swinver1, tnt, wang2021pvtv2}. The second approach involves using hybrid architectures, such as introducing convolutional blocks within ViTs~\cite{cvt, how_vt}. However, both approaches primarily benefit medium-sized datasets and not small-scale datasets.
Several efforts have been made to enhance their locality inductive bias, as reported in literature~\cite{efficient, local, small_data_wacv, smalldata}. 
Among these methods, SSL has demonstrated exceptional efficacy in training transformers from scratch on small datasets~\cite{smalldata, efficient, cdnet_med, hipt, transpath, chen2022self, scorenet}.  
These methods typically involve sequentially conducting SSL and fine-tuning on the same small dataset to enhance ViT performance. 

Meanwhile, another straightforward approach that takes advantage of SSL is to jointly optimize the self-supervised task along with the primary task like classification or segmentation. 
We name such SSL tasks as \textbf{S}elf-\textbf{S}upervised \textbf{A}uxiliary \textbf{T}ask (\textbf{SSAT}).
Although SSAT has been explored in the vision community~\cite{efficient, local, evolving_losses} and robotics community~\cite{laskin2020curl, lidoes}, there are still many open questions, especially when the size of the dataset is limited.

This paper empirically analyzes the aforementioned joint learning approach with SSAT, as an alternative to sequentially performing SSL and fine-tuning (SSL+FT) on the same dataset.
Through an extensive amount of experiments on \textit{ten} image classification datasets of various sizes as well as \textit{two} video classification datasets, surprisingly, we observe that SSAT works significantly better than other baselines like SSL+FT and training from scratch, especially for ViT on small datasets (see Figure~\ref{fig:intro}).
Further experiments empirically show that it is most effective when the auxiliary task is image reconstruction from missing pixels among the well known SSL methods we tested.
Finally, we perform a detailed model and feature analysis to highlight the unique properties of SSAT-driven models in comparison to other representative baselines. This distinction is particularly notable when comparing with the SSL+FT models which are trained with similar loss functions.
We reveal that the advantages of SSAT in a limited-data regime come from better semantic richness, a distinct attention distribution, and an increased capability for feature transformation, which results in higher feature variance.

\begin{figure*}
    \centering
    \scalebox{0.8}{
    \includegraphics[width=.95\linewidth]{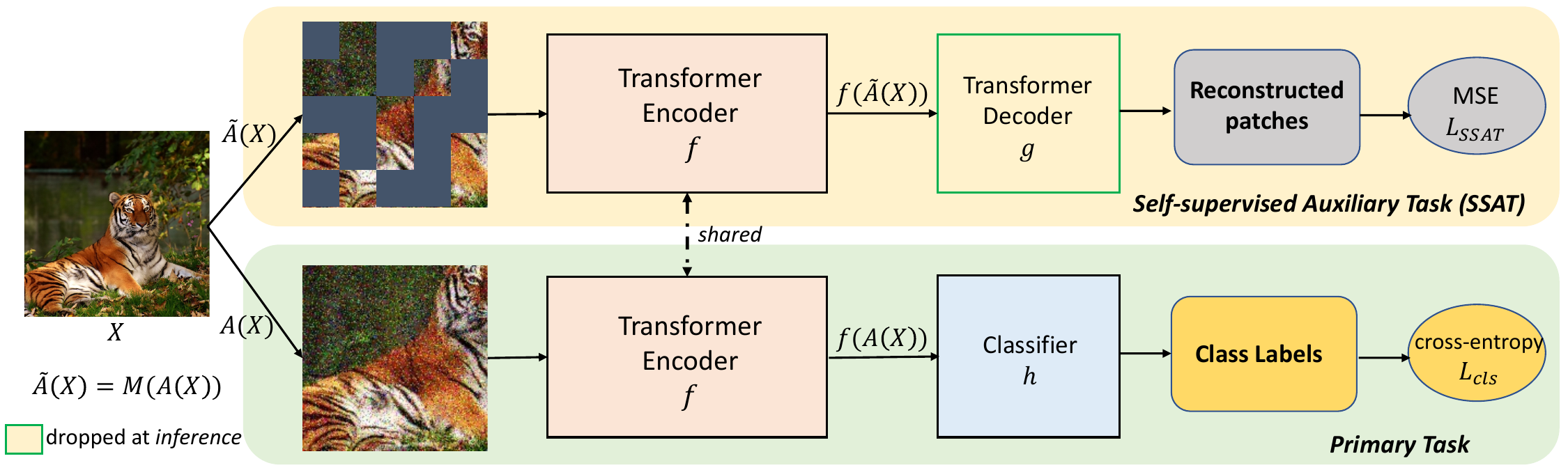}}
    \caption{An overview of \textbf{ViT training with SSAT}. The input $X$ to the ViT undergoes data augmentations, $A(X)$ and $\Tilde{A}(X) = M(A(X))$, using a mask operation M. These augmented inputs are then fed to a Transformer Encoder $f$, resulting in two latent representations: $f(A(X))$ and $f(\Tilde{A}(X))$. These correspond to the masked and full image, respectively. The latent representation of the full image is utilized for the image classification task, while the masked image's representation is used for the image reconstruction task. ViT training involves joint optimization of losses from both these tasks.} \vspace{-0.1in}
\label{fig:framework}
\end{figure*}

\section{Related Work}

\textbf{Vision Transformers.} Several vision transformers~\cite{zhu2020deformable,dosovitskiy2020vit,zhou2021deepvit,wang2021pvtv2,xie2021segformer,cheng2021per,arnab2021vivit,liu2021video,timesformer, ryoo2021tokenlearner, ranasinghe2022self} have been introduced in recent times for a wide range of tasks. However, these models require large-scale pre-training to be effective on different datasets.
In an effort to reduce their reliance on extensive training, DeiT~\cite{deit} introduced extensive data augmentation, regularization, and distillation tokens from convolutions in ViTs. T2T~\cite{t2t}, in a similar vein, employed a tokenization technique that flattened overlapping patches and applied a transformer to allow for learning local structural information around a token. Meanwhile, some ViT models~\cite{heo2021pit, cvt, dai2022mstct} have introduced inductive bias into the transformers through the use of convolutional filters. Hierarchical transformers~\cite{swinver1, mvit1, mvit2, wang2021pyramid} have introduced inductive bias by reducing the number of tokens through patch merging and thus operating at different scales. However, these architectures do not overcome the limitation of ViTs, which require at least a medium-sized dataset for pre-training~\cite{how_vt}.


\textbf{Self-supervised Learning.}
Self-Supervised Learning (SSL) aims to learn visual representations through pretext tasks. Contrastive methods, such as SimCLR~\cite{simclr} and MoCo~\cite{moco}, minimize the distance between differently augmented views of the same image (positive pairs) while maximizing it for dissimilar images (negative pairs). On the other hand, non-contrastive methods like BYOL~\cite{byol} and DINO~\cite{dino} only impose minimization between the positive pairs. In contrast, reconstruction based methods~\cite{mae, convmae, simmim, videomae} have shown to be effective self-supervised learners for various downstream computer vision tasks. In these methods, an encoder operates on a small portion of an image to learn a latent representation, and a decoder decodes the latent representation to reconstruct the original image in the pixel space.
These SSL methods are commonly used for large-scale pre-training of ViTs to enhance their effectiveness in various downstream tasks.


\textbf{ViTs for small datasets.}
    Liu et al.~\cite{efficient} proposed an auxiliary self-supervised task that improves the robustness of ViT training on smaller datasets. The task involves predicting relative distances among tokens and is jointly trained with primary tasks. On the other hand, Li et al.\cite{local} conducted distillation in the hidden layers of ViT from a lightweight CNN-trained model.
    To address the lack of locality inductive bias, Lee et al.\cite{small_data_wacv} introduced a ViT architecture with shifted patch tokenization and locality self-attention. 
    Gani et al.~\cite{smalldata} proposed an SSL+Fine-tuning methodology where the SSL is similar to the pretext task in DINO~\cite{dino}. These methods eliminate the need for large-scale pre-training and allow ViTs to learn meaningful representations with limited data. In contrast to these methods, we propose SSAT akin to~\cite{efficient}, but with an approach that combines the functionality of self-attention and MLPs through image reconstruction.

\vspace{-2mm}

\section{Preliminaries}
\label{sec:vitprelims}
ViT utilizes a non-overlapping grid of image patches to process a given image $X$, where each patch is linearly projected into a set of input tokens. ViT consists of a stack of multi-head attention and linear layers as in~\cite{dosovitskiy2020vit}. 
The transformer attention layers model the pairwise relationship between the input tokens~\cite{attention_is_all_you_need}. For generalizability, We denote the transformer encoder as $f$. For brevity, we have omitted the parameters of the encoder. In practice, $f$ operates on an augmented version of the input image $X$ to output a discriminative representation  $f(A(X))$ where $A$ is the set of image augmentation. This representation is subsequently classified into class labels using a classifier $h$. A class-wise cross-entropy loss $L_{cls}$ is used to train the transformer encoder.

\section{Self-supervised Auxiliary Task (SSAT)}
Our objective is to improve the ViT training on the dataset with limited samples. Consequently, we propose to jointly train the primary classification task of ViT alongside a self-supervised auxiliary task (SSAT). The joint optimization of the SSAT and classification task allows the ViT to capture inductive biases from the data without requiring any additional labels. An overview of our framework is depicted in Figure \ref{fig:framework}.

In our joint optimization framework for SSAT, we have utilized the widely adopted Masked Autoencoder (MAE) approach~\cite{MaskedAutoencoders2021} for reconstructing the missing pixels. Nonetheless, it is worth noting that any SSL method can be integrated into our framework, given its generic nature. Our decision to use MAE was based on its superior performance, as evidenced by our experimental analysis (Table~\ref{joint_ssl}). 

To the existing ViT frameworks, where the Transformer encoder $f$ and Classifier $h$ process the full image patches $A(X)$ to compute the classification loss $L_{cls}$.
we introduce an augmentation set $\Tilde{A} = M(A(X))$, where operation $M$ randomly masks out patches in the input image $X$.
The transformer encoder $f$ also operates on the unmasked tokens, generating latent representation $f(\Tilde{A}(x))$ for these tokens. 
In parallel to the classifier $h$ 
, SSAT employs a shallow decoder $g$ to reconstruct back the unseen image pixels
from the latent representation of the seen tokens $f(\Tilde{A}(x))$. 
Following~\cite{MaskedAutoencoders2021}, the decoder takes as input the latent representation of the seen tokens $f(\Tilde{A}(x))$ and a learnable masked token. 
Each token representation at the decoder's output is linearly projected to a vector of pixel values representing a patch. 
The output $g(f(\Tilde{A}(X)))$ is reshaped to form the reconstructed image, thereafter computing the normalized Mean Square Error (MSE) loss $L_{SSAT}$ between the original and reconstructed image. In practice, the MSE is computed only for the masked patches as in~\cite{mae}. 

Thus, the entire framework performs a primary task, i.e. \textit{classification} and a self-supervised auxiliary task, i.e. \textit{reconstruction}.
This framework can be jointly optimized using a convex combination of the losses from the primary task and SSAT. Thus, the total loss is computed by

\begin{equation}
    L = \lambda * L_{cls} + (1-\lambda) * L_{SSAT}
\end{equation}
$\lambda$ is the loss scaling factor. 
During inference, the decoder is discarded and the encoder $f$ processes all input patches to generate the classification output only.
Our framework supports training of any ViT model and SSAT variants. 
\begin{table*}[h!]
\tiny
    \caption{Top-1 classification accuracy (\%) of different ViT variants with and without SSAT on CIFAR-10, CIFAR-100, Flowers102, and SVHN datasets. All models were trained for 100 epochs.}
    \centering
    \tiny
    \scalebox{0.9}{
    \resizebox{0.9\textwidth}{!}{\begin{tabular}{c | c | c c c c}
    \hline
    \textbf{Method} & \textbf{\# params. (M)} & \textbf{CIFAR-10} & \textbf{CIFAR-100} & \textbf{Flowers102} & \textbf{SVHN}\\
    \hline
    ViT-T~\cite{deit} & 5.4 & 79.47 & 55.11 & 45.41 & 92.04\\
    \hspace{0.2cm}\textbf{+SSAT} & 5.8 & \textbf{91.65} (\textcolor{green}{+12.18}) & \textbf{69.64} (\textcolor{green}{+14.53}) & \textbf{57.2} (\textcolor{green}{+11.79}) & \textbf{97.52} (\textcolor{green}{+5.48}) \\ \hline
    
    ViT-S~\cite{deit} & 21.4 & 79.93 & 54.08 & 56.17 & 94.45 \\
    \hspace{0.2cm}\textbf{+SSAT} & 21.8 & \textbf{94.05} (\textcolor{green}{+14.12}) & \textbf{73.37}   (\textcolor{green}{+19.29}) & \textbf{61.15} (\textcolor{green}{+4.98}) & \textbf{97.87} (\textcolor{green}{+3.42})  \\ \hline
    
    CVT-13~\cite{cvt}  & 20 & 89.02 & 73.50 & 54.29 & 91.47 \\
    \hspace{0.2cm}\textbf{+SSAT} & 20.3 & \textbf{95.93} (\textcolor{green}{+6.91}) & \textbf{75.16} (\textcolor{green}{+1.66}) & \textbf{68.82} (\textcolor{green}{+14.53}) & \textbf{97} (\textcolor{green}{+5.53}) \\  \hline
    
    Swin-T~\cite{swinver1}  & 29 & 59.47 &  53.28 &  34.51 & 71.60 \\
    \hspace{0.2cm}\textbf{+SSAT}  & 29.3 & \textbf{83.12} (\textcolor{green}{+23.65}) & \textbf{60.68} (\textcolor{green}{+7.4})  & \textbf{54.72} (\textcolor{green}{+20.21}) & \textbf{85.83} (\textcolor{green}{+14.23}) \\ \hline
    
    ResNet-50~\cite{resnet-50} & 25.6 & 91.78 & 72.80 & 46.92 & 96.45 \\
    \hline
    \end{tabular}}}
    \label{compare}
\end{table*}
\section{Experimental Analysis}
\label{sec:formatting}
In this section, we present the superiority of using SSAT while training any vision transformer. Our experiments are based on image and video classification tasks. We use 12 different datasets: (i) 4 small sized datasets: CIFAR-10~\cite{cifar}, CIFAR-100~\cite{cifar}, Oxford Flowers102~\cite{flowers} (Flowers) and SVHN~\cite{svhn}, (ii) 1 medium sized dataset: ImageNet-1K~\cite{imagenet_cvpr09} (IN-1K), (iii) 2 medical datasets: Chaoyang~\cite{chaoyang} and PMNIST~\cite{medmnist}, (iv) 3 datasets of DomainNet~\cite{domainnet}: ClipArt, Infograph, and Sketch, and (v) 2 video datasets for deepfake detection: DFDC~\cite{seferbekov2020dfdc} and FaceForensics++~\cite{roessler2019faceforensicspp}. 

Our experiments for image reconstruction in the context of SSAT generally follow the procedure outlined in~\cite{mae}, unless otherwise stated. In particular, we employ the decoder design from~\cite{mae} for ViT and utilize the decoder design from ConvMAE~\cite{convmae} and SimMIM~\cite{simmim} for hierarchical encoders such as CVT and swin, respectively. To optimize hyper-parameters for the decoder, we conduct our experiments with the ViT encoder. For augmentation $\Tilde{A}$, we use a random masking with 75\% masking ratio. Our decoder has a depth of 2 (i.e. 2 transformer layers) and an embedding dimension of 128. We provide ablations on the choice of these hyper-parameters in Appendix~\ref{decoder_appendix}. It is worth noting that our decoder is shallower than that in MAE~\cite{mae}. The loss scaling factor $\lambda$ is set to $0.1$ for all the datasets. 

Our ViT encoders ($f$) are trained using the training recipe of DeiT~\cite{deit}, unless otherwise specified. The configuration of ViT-T, ViT-S, and ViT-B is identical to the configuration described in~\cite{deit}. We borrow the network architecture for CVT-13, and Swin from the official code of \cite{cvt}, and \cite{swinver1}, respectively. Training is conducted for 100 epochs, unless otherwise specified, using 8 A5000 24GB GPUs for IN-1K and one A5000 24 GB GPU for all other datasets. Additional training details for each dataset can be found in Appendix~\ref{training_appendix}.

\subsection{Main Results}

\noindent \textbf{SSAT on small-sized dataset}: In Table~\ref{compare}, we present the classification accuracy on the small-sized datasets with different variants of vision transformers: ViT-T, ViT-S, CVT-13, and Swin-T. In this table, we demonstrate the impact of using SSAT while training the transformers for learning the class labels. All the models have been trained for 100 epochs from scratch. Although the models with SSAT have more training parameters, they have identical operations during inference. SSAT improves the classification accuracy on all the datasets for all the transformer encoders. It is worth noting that ViT-T with 5.4M parameters when trained with SSAT outperforms ViT-S with 21.4M parameters. The highest classification accuracy is achieved with CVT-13 (20M parameters) due to the introduction of convolutions that infuse inductive bias into the transformers. Although convolutions are generally more effective than transformers on small datasets, our experiments demonstrate that the most effective convolutional network (ResNet-50~\cite{resnet-50}) for these datasets underperforms most of the transformers when trained with SSAT, except for ViT-T on CIFAR-10 and CIFAR-100 datasets.

\begin{table*}[!htb]
    \parbox{.4\linewidth}{
    \begin{minipage}[t]{.4\textwidth}
    \centering
    \caption{Top-1 classification accuracy (\%) on ImageNet-1K (IN-1K), perturbed CIFAR-100 (CIFAR-100-\textit{p}), and perturbed ImageNet-1K (IN1K-\textit{p})}
    \label{perturbed_imagenet_cifar}
    \scalebox{0.85}{
    \begin{tabular}{ l |c c c } 
         \hline
          \textbf{Method} & \textbf{IN-1K} & \textbf{CIFAR-100-\textit{p}} & \textbf{IN-1K-\textit{p}} \\
         \hline
         ViT-T & 65.0 & 25.1 & 48.3 \\
         \rowcolor{Gray}
         \hspace{0.2cm}\textbf{+SSAT} & \textbf{72.7 } & \textbf{37.6 } & \textbf{59.6 } \\
         \hline
         ViT-S & 74.2& 22.5  & 62.7 \\
         \rowcolor{Gray}
         \hspace{0.2cm}\textbf{+SSAT} & \textbf{76.4} & \textbf{43.9 }  & \textbf{64.5} \\
         \hline
    \end{tabular}}
    \end{minipage}
    }
    \hfill
    \parbox{.575\linewidth}{
    \begin{minipage}[t]{.575\textwidth}
    \centering
    \caption{Top-1 accuracy and efficiency of ViT-T trained from scratch, with SSL+FT, and with SSAT. We provide the GFLOPs, training time (GPU hours), and CO$_2$ emissions (kg eq) for IN-1K.}
    \label{sslft}
    \scalebox{0.7}{
    \begin{tabular}{c | c| c c c |c|c}
        \hline
        \multirow{2}{*}{\textbf{Method}} & \multirow{2}{*}{\textbf{GFLOPs}} & \multirow{2}{*}{\textbf{CIFAR-10}} & \multirow{2}{*}{\textbf{CIFAR-100}} & \multirow{2}{*}{\textbf{IN-1K}} & \small{\textbf{Train}} & \textbf{Kg CO$_2$} \\
         &  &  & & & \small{\textbf{time}} & \textbf{eq.} \\
        \hline
        Scratch & 1.26 & 79.47 & 55.11 & 65.0 & 60 & 5.96 \\
        \small{(1) SSL+FT} & 0.43+1.26 & 85.33 & 60.43 & 70.09 & 55 & 5.46 \\
        \small{(2) SSL+FT} & 0.43+1.26 & 86.48 & 63.28 & 71.1 & 82 & 8.15 \\
        \small{(3) SSL+FT} & 0.43+1.26  & 85.3 & 60.3 & 70.5 & 74 & 7.35 \\ \hline
        \small{(4) SSL+FT} & 0.43+1.26 & 88.72 & 67.53 & \textbf{74.07} & 104 & 10.33 \\
        \rowcolor{Gray}
        Ours  & 1.67 & \textbf{91.65} & \textbf{69.64} & 72.69 & 78 & 7.55 \\
        \hline
    \end{tabular}}
    
    \end{minipage}}
\end{table*}

\noindent \textbf{SSAT on medium-sized dataset}:   In Table~\ref{perturbed_imagenet_cifar}, we present the impact of SSAT on ViTs that were trained on a medium-sized dataset, such as IN-1K~\cite{imagenet_cvpr09}. Our results demonstrate that SSAT consistently enhances the classification accuracy of ViTs, even as the number of training samples increases. Notably, this improvement is more pronounced for smaller models, which have 5.4M parameters, than for larger ones. Specifically, we observed a relative performance improvement of 11.8\% for ViT-T with SSAT, as compared to only 2.9\% for ViT-S+SSAT. These findings suggest that SSAT can be effectively utilized to train lighter transformers that can be deployed on edge devices.

\noindent \textbf{Does SSAT promotes overfitting?}  In Table~\ref{perturbed_imagenet_cifar}, we also analyse the robustness of ViTs to natural corruptions. Given that we recommend the use of SSAT to enhance representation learning in transformer training, it is reasonable to question whether this approach can lead to overfitting on small training samples. To address this concern, we evaluate the performance of our trained models on perturbed versions of the data, specifically, CIFAR-100-p and IN-1K-p, which are obtained by applying random perspective transformations to images following~\cite{3dtrl}. Our results demonstrate that ViTs trained with SSAT exhibit greater robustness to these natural corruptions compared to the baseline ViTs. We observe notable improvements in performance for tiny ViTs, as evidenced by the results for ViT-T+SSAT in Table~\ref{perturbed_imagenet_cifar}, as well as for smaller datasets such as CIFAR-100-p.

\begin{table}
    \centering
    \caption{Top-1 accuracy of existing SSL strategies used as SSAT. MAE as the SSAT achieves the best result on both CIFAR-10 and CIFAR-100.}
    \label{joint_ssl}
    \scalebox{0.75}{
    \begin{tabular}{c|c|c}
    \toprule
    \textbf{SSAT (SSL)} & \textbf{CIFAR-10} & \textbf{CIFAR-100}\\ \midrule
    SimCLR~\cite{simclr} & 55.21 & 36.49\\
    DINO~\cite{dino} & 80.07 & 60.6\\
    MAE~\cite{mae} & \textbf{91.65} & \textbf{69.64}\\
    \bottomrule
    \end{tabular}}
    \vspace{-15pt}
\end{table}

\noindent \textbf{Comparison of SSAT with SSL+FT}: In Table~\ref{sslft}, we present the superiority of joint training of the SSL loss with the classification loss over the two-step sequential training approach, where the model is first trained with SSL and then fine-tuned (FT) for classification. Our empirical analysis is conducted on ViT-T, where we compare the performance of ViT trained from scratch and ViT + SSAT, which are trained for 100 epochs. To establish baselines for our SSL+FT model, we conducted experiments using four different training protocols: (1) 50 epochs of SSL training followed by 50 epochs of fine-tuning, (2) 50 epochs of SSL training followed by 100 epochs of fine-tuning, (2) 100 epochs of SSL training followed by 50 epochs of fine-tuning, and (4) 100 epochs of SSL training followed by 100 epochs of fine-tuning. Additionally, we quantify the carbon emission of the models trained using different methods with the help of a tool provided by~\cite{co_emmission}. Note that the GFLOPs, training time, and Kg CO$_2$ eq. are specified for the model trained on IN-1K for better generalizability. Our empirical results show that all models incorporating SSL outperform those trained from scratch, highlighting the importance of self-supervised learning when training transformers on small datasets. Moreover, even when requiring an additional 4 hours of training time and resulting in approximately 0.6 Kg CO$_2$ equivalent of additional carbon emissions, our SSAT models demonstrate superior performance compared to the SSL+FT model (50 epoch SSL + 100 epoch FT).
Although accuracy improves when SSL+FT models are trained on CIFAR-10 and CIFAR-100 for 104 GPU hours, our SSAT approach remains superior, requiring 26 GPU hours less training time and burning approximately 2.8 Kg CO$_2$ equivalent. However, the SSL+FT model outperforms SSAT when a large amount of training data is available.

\noindent \textbf{Appropriate SSL for joint training}: Table~\ref{joint_ssl} presents a comparison of the performance of the SSAT approach, implemented with different SSL strategies, namely, contrastive (SimCLR~\cite{simclr}), non-contrastive (DINO~\cite{dino}), and reconstruction based (MAE~\cite{mae}), on the ViT model. Our analysis reveals that the use of SimCLR results in a decrease in the ViT's performance, which can be attributed to the conflicting losses that arise while optimizing the cross-entropy loss to learn class labels and the contrastive loss. However, DINO and MAE both enhance the ViT's performance when jointly trained with cross-entropy. Notably, the improvement observed with MAE is more significant than that with DINO. The superior performance of MAE can be attributed to the centering and sharpening technique employed in DINO, which impedes the learning of class labels while only facilitating the SSL. On the other hand, as mentioned in~\cite{ssl_learn}, MAE encourages MLPs in ViTs to be more representative. While the cross-entropy loss primarily contributes more to the self-attention blocks. Thus, SSAT implemented with reconstruction based SSL harmonizes the impact of both tasks, thus improving the ViT's learning capabilities.

\begin{table}[h]
    \centering
    \caption{Top-1 accuracy on medical image datasets. All models are trained for 100 epochs.}
    \label{medical}
    \scalebox{0.9}{
    \begin{tabular}{cc|cc}
    \hline
    & \textbf{Method} & \textbf{Chaoyang} & \textbf{PMNIST}\\
    \hline
    \multirow{3}{*}{\rotatebox{90}{ViT-T}}& Scratch  & 77.37 & 90.22\\
    & IN-1K pretrained + FT  & 78.78 & 91.99\\ 
    & Scratch + \textbf{SSAT}  & \textbf{82.52} & \textbf{93.11}\\
    \hline
    \multirow{3}{*}{\rotatebox{90}{ViT-S}} & Scratch  & 80.04 & 91.19 \\
    & IN-1K pretrained + FT & 80.18 & 92.63 \\ 
    & Scratch + \textbf{SSAT}  & \textbf{81.25} & \textbf{93.27}\\
    \hline
    \end{tabular}}
\end{table}

\begin{table}
    \centering
    \caption{Top-1 accuracy on DomainNet datasets. All models are trained for 100 epochs}
    \label{DomainNET}
    \scalebox{0.9}{
    \begin{tabular}{c| c c c }
    \hline
    \textbf{Method} & \textbf{ClipArt} & \textbf{Infograph} & \textbf{Sketch}\\
    \hline
    ViT-T & 29.66 & 11.77 & 18.95\\
    \rowcolor{Gray}
     \hspace{0.2cm}\textbf{+SSAT} & 47.95 & 16.37 & 46.22\\
     \hline
     CVT-13 & 60.34 & 19.39 & 56.98 \\
      \hspace{0.2cm}+$\mathcal{L}_{drloc}$~\cite{efficient} & 60.64 & 20.05 &  57.56\\
      \rowcolor{Gray}
    \hspace{0.2cm}\textbf{+SSAT} & \textbf{60.66} & \textbf{21.27} & \textbf{57.71}\\
    \hline
    \end{tabular}}
    \vspace{-15pt}
\end{table}

\begin{figure*}
\centering
\begin{minipage}{.3\textwidth}
  \centering
  \includegraphics[width=0.85\linewidth]{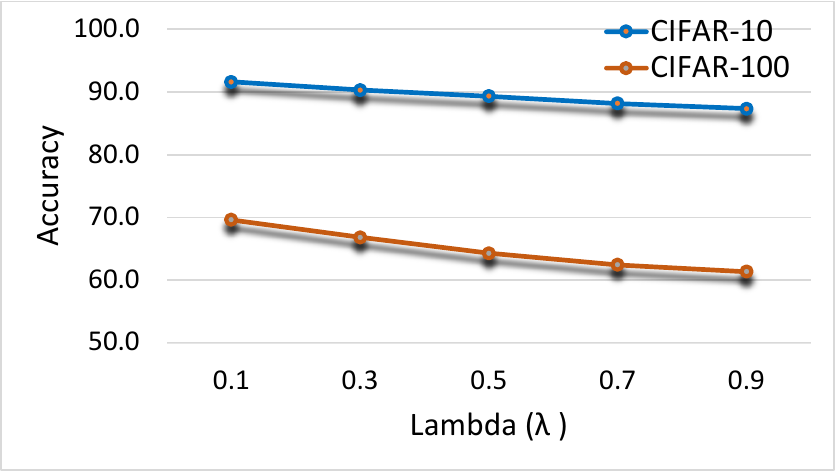}
  \vspace*{-7pt}
  \captionof{figure}{Ablation for loss scaling\\ factor $\lambda$.}
  \label{fig:test1}
\end{minipage}%
\begin{minipage}{.3\textwidth}
  \centering
  \includegraphics[width=0.85\linewidth]{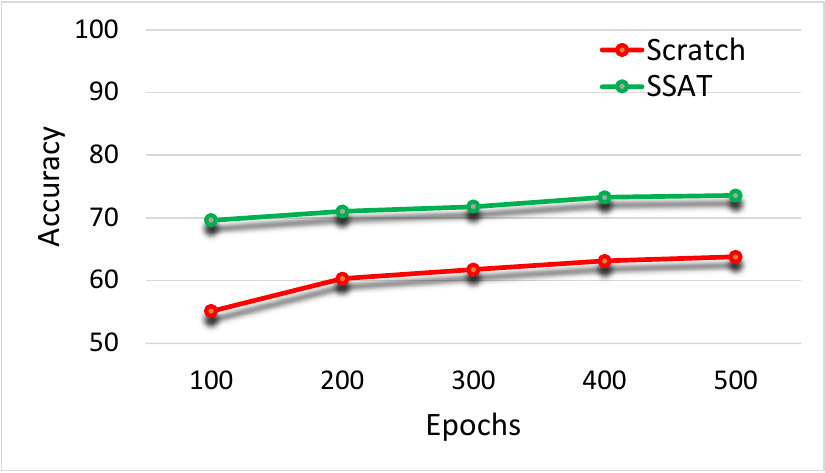}
  \vspace*{-7pt}
  \captionof{figure}{ViT (scratch) vs. ViT+SSAT \\ for longer epochs on CIFAR-100.}
  \label{fig:test2}
\end{minipage}%
\begin{minipage}{.3\textwidth}
  \centering
  \includegraphics[width=0.85\linewidth]{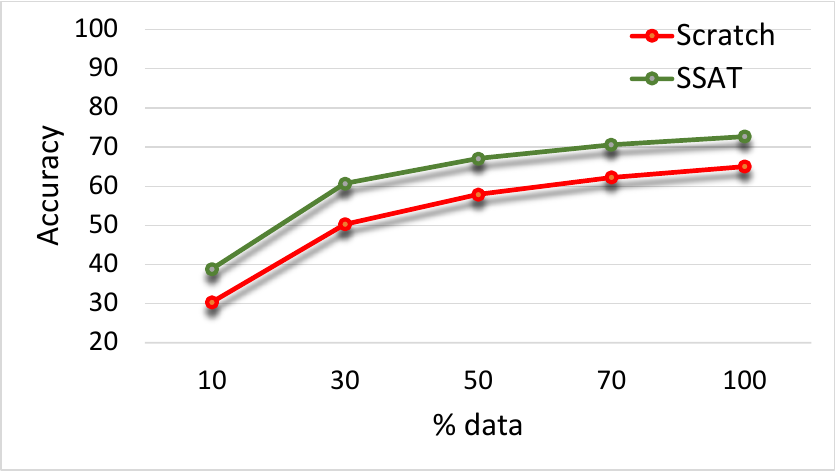}
  \vspace*{-7pt}
  \captionof{figure}{ViT (scratch) vs. ViT+ \\ SSAT for different subset of IN-1K.}
  \label{fig:test3}
\end{minipage}
    \vspace{-15pt}
\end{figure*}

\noindent \textbf{Superiority of SSAT over Large-scale pre-training}: In situations where training samples are limited and the data distribution differs from that of natural images, large-scale pretraining can be challenging. The main obstacle is the lack of data that accurately represents the downstream data distribution. Consequently, we conducted experiments using ViTs on medical and domain adaptation datasets (Tables~\ref{medical} and~\ref{DomainNET}) where data is scarce. In Table~\ref{medical}, we demonstrate how SSAT significantly enhances the classification performance of ViT-T on the Chaoyang and PMNIST datasets. The resulting model not only surpasses a comparable ViT model that was pre-trained on ImageNet~\cite{imagenet_cvpr09}, but also outperforms its larger ViT-S model when trained without SSAT.  
We observed similar trends of improvement on three datasets from DomainNet~\cite{domainnet} in Table~\ref{DomainNET}. It is worth mentioning that our CVT model, when trained using SSAT, outperforms $\mathcal{L}_{drloc}$~\cite{efficient}, which is another state-of-the-art self-supervised loss designed to enhance transformer performance on small datasets.

\noindent \textbf{Loss scaling factor}: In Figure~\ref{fig:test1} we perform an empirical analysis to determine the optimal value for the loss scaling factor $\lambda$. Our analysis focused on CIFAR datasets show that the choice of $\lambda = 0.1$ is an optimal choice when SSAT positively impacts the primary classification task.

\noindent \textbf{Extended training}: In this experiment, we extend the training schedules of both the scratch and SSAT model as illustrated in Figure~\ref{fig:test2}. Our findings indicate that the performance enhancement of our SSAT model, relative to the ViT baseline, remains consistent throughout the entire training period. These results suggest that the improvement in the SSAT model's performance is not due to a faster convergence rate, but rather to superior optimization capabilities.

\noindent \textbf{Training for different subsets of IN-1K}: Figure~\ref{fig:test3} presents our analysis of the performance of the ViT baseline and SSAT model for varying training sample sizes, specifically on subsets of IN-1K. Our results demonstrate that the performance enhancement of the SSAT model, relative to the baseline model, is consistent across all subsets (i.e., different sizes of the training data). These findings substantiate that models with low training parameters, such as ViT-T, can benefit from SSAT at all scales of training data.

\begin{figure*}[h]
     \centering
     \begin{subfigure}[b]{0.25\textwidth}
         \centering
         \includegraphics[width=\textwidth]{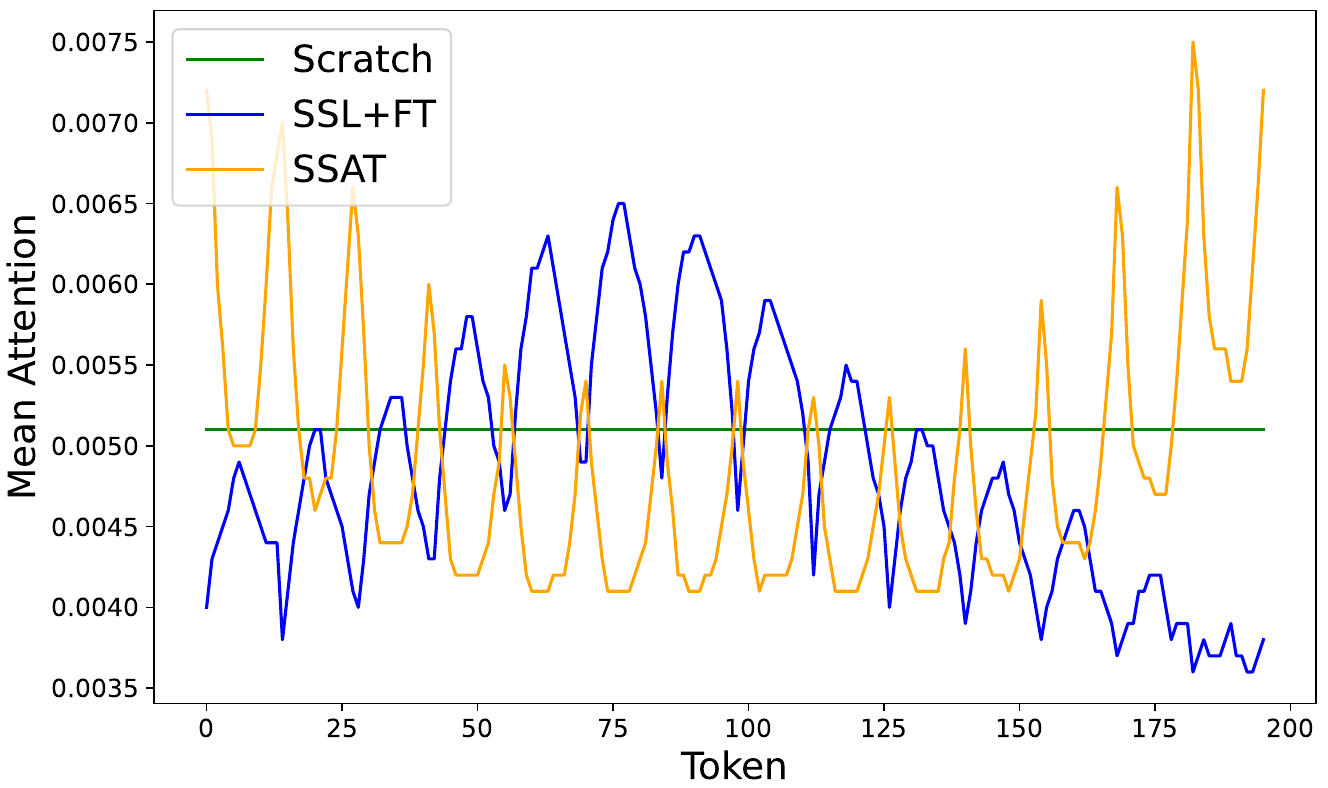}
     \end{subfigure}
     \begin{subfigure}[b]{0.25\textwidth}
         \centering
         \includegraphics[width=\textwidth]{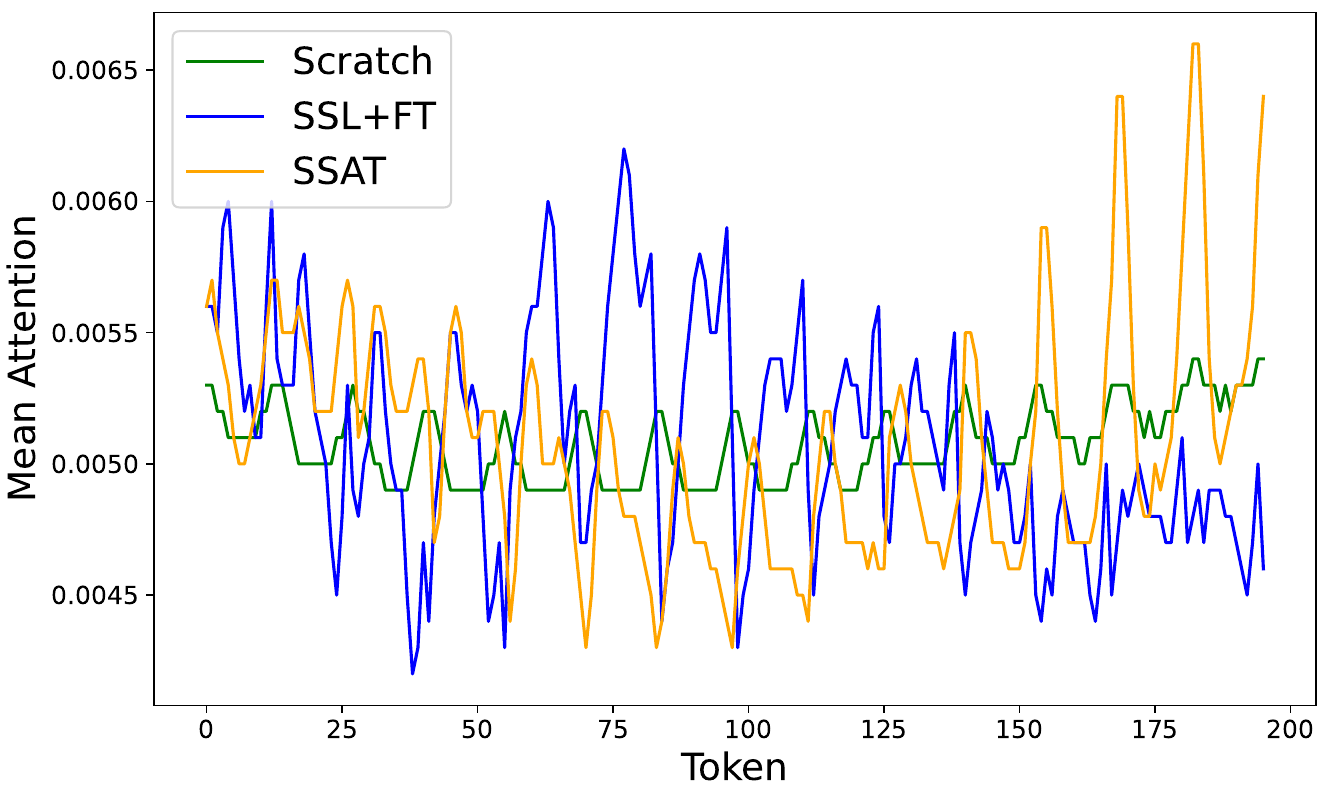}
     \end{subfigure}
     \begin{subfigure}[b]{0.25\textwidth}
         \centering
         \includegraphics[width=\textwidth]{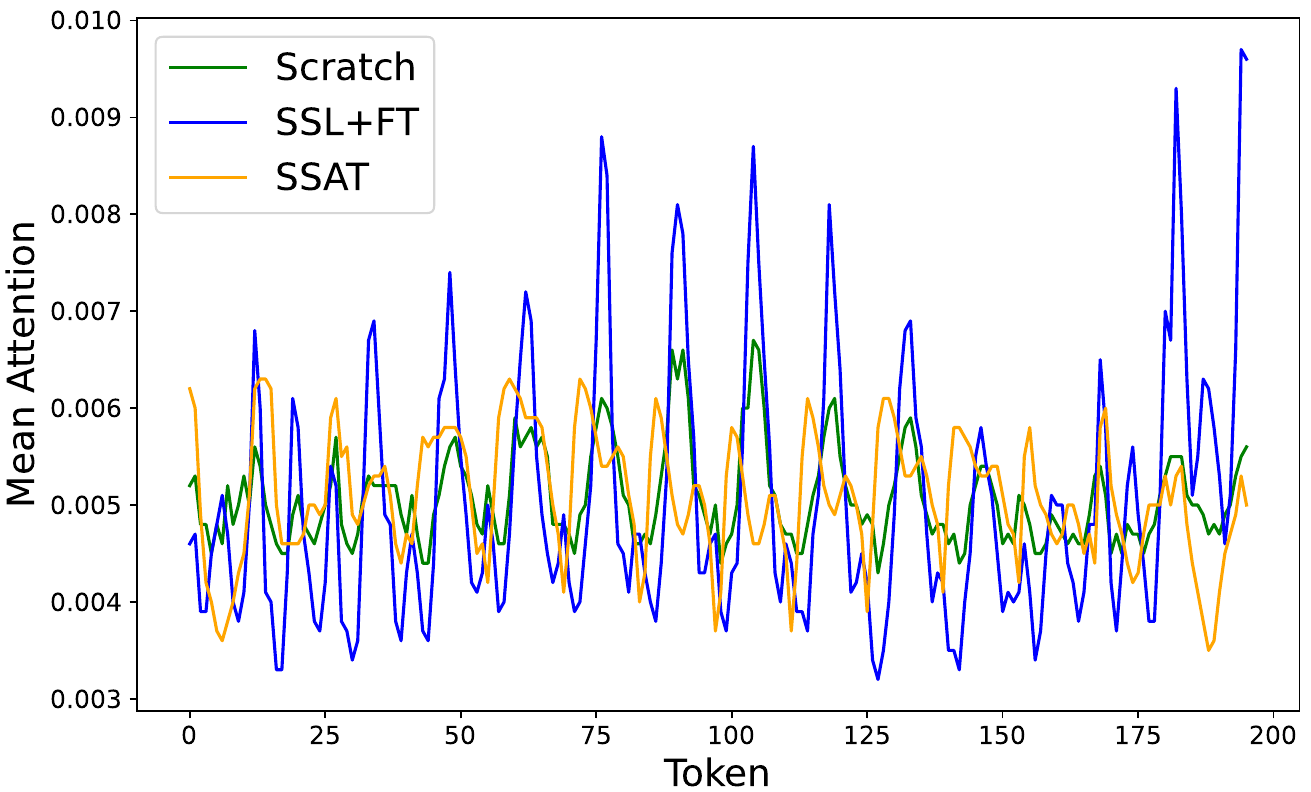}
     \end{subfigure}
     \begin{subfigure}[b]{0.25\textwidth}
         \centering
         \includegraphics[width=\textwidth]{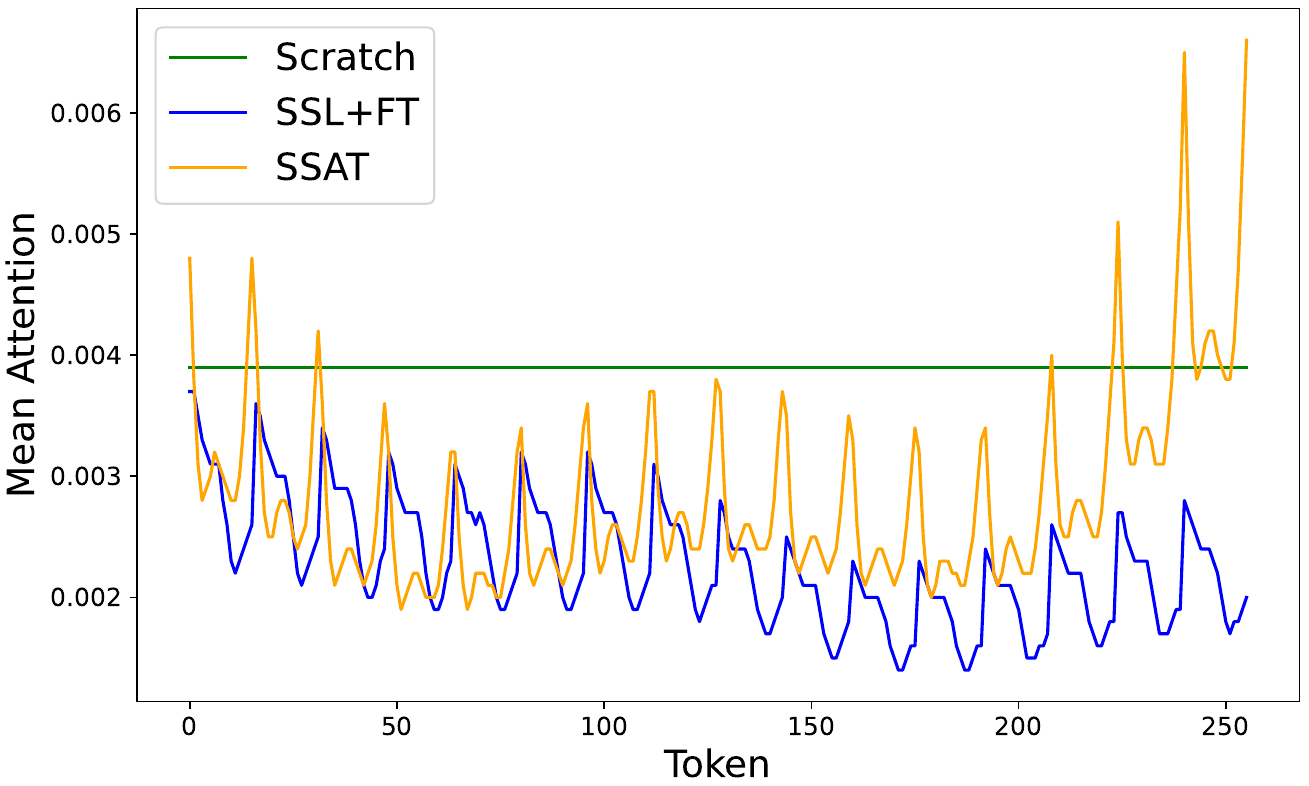}
     \end{subfigure}
     \begin{subfigure}[b]{0.25\textwidth}
         \centering
         \includegraphics[width=\textwidth]{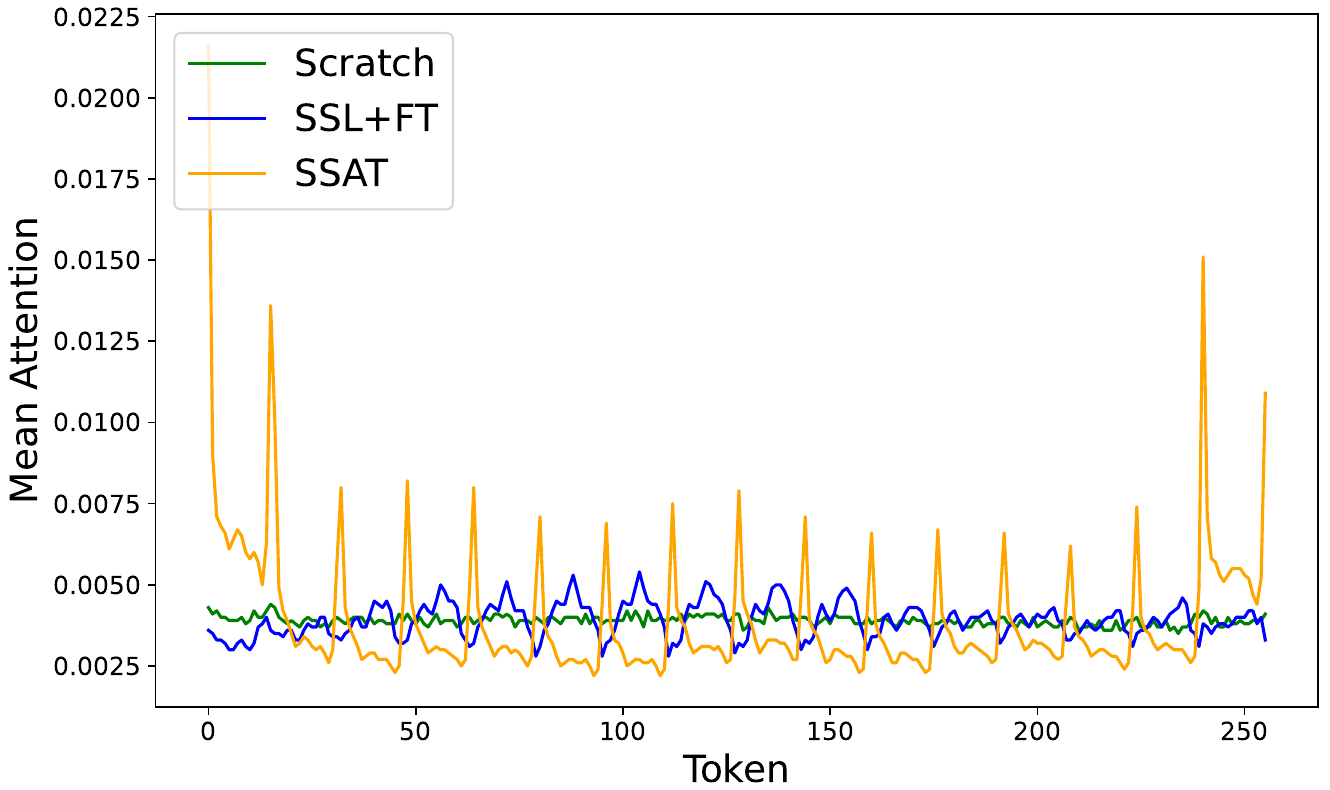}
     \end{subfigure}
     \begin{subfigure}[b]{0.25\textwidth}
         \centering
         \includegraphics[width=\textwidth]{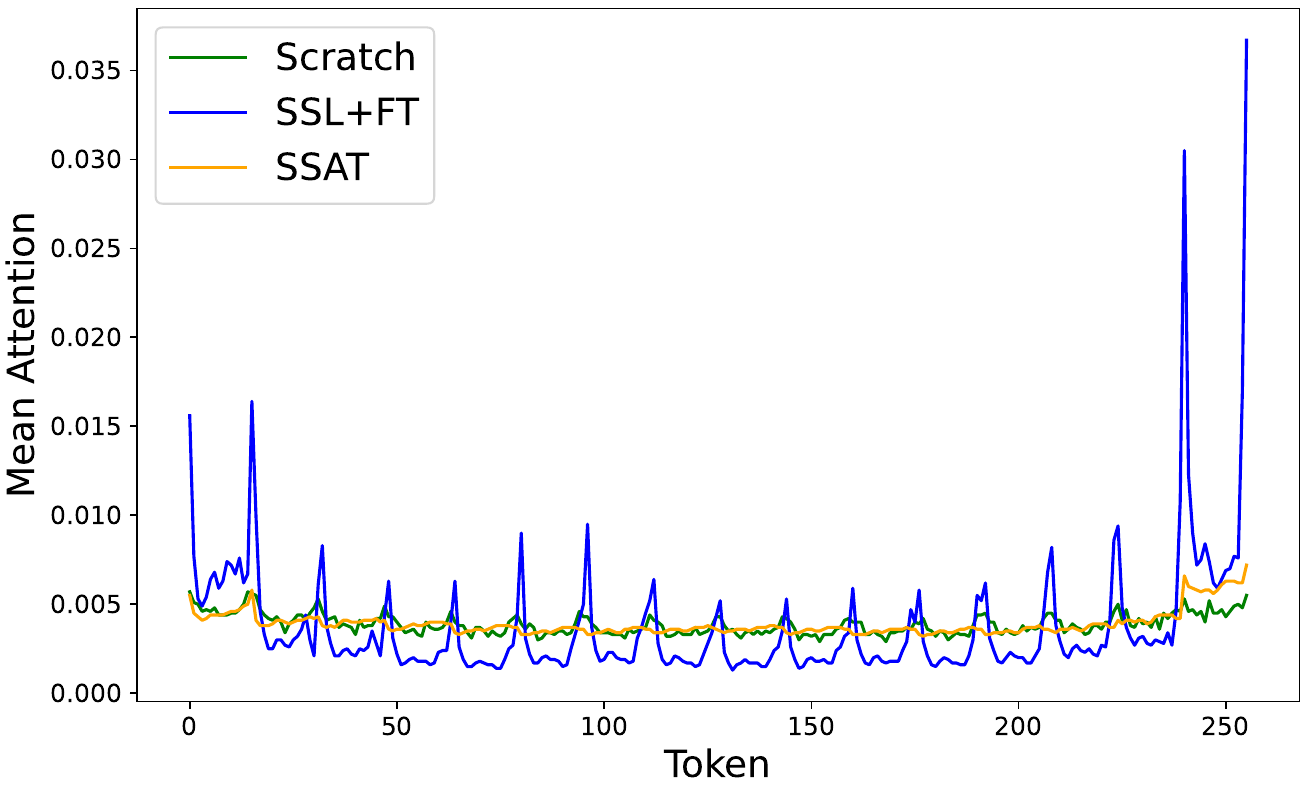}
     \end{subfigure}
    \caption{The \textbf{distribution of attention weights} across the $n$ tokens for different ViT-T blocks on two datasets: Oxford Flower (top row) and CIFAR-100 (bottom row). The first, second, and third columns correspond to the attention distributions of the first, sixth, and twelfth ViT-T blocks, respectively.} \vspace{-0.1in}
    \label{flower:bin-plots}
\end{figure*}
\begin{figure}[h]
     \centering
     \graphicspath{{./images/}}
     \begin{subfigure}[h]{0.2\textwidth}
         \centering
         \includegraphics[width=\textwidth]{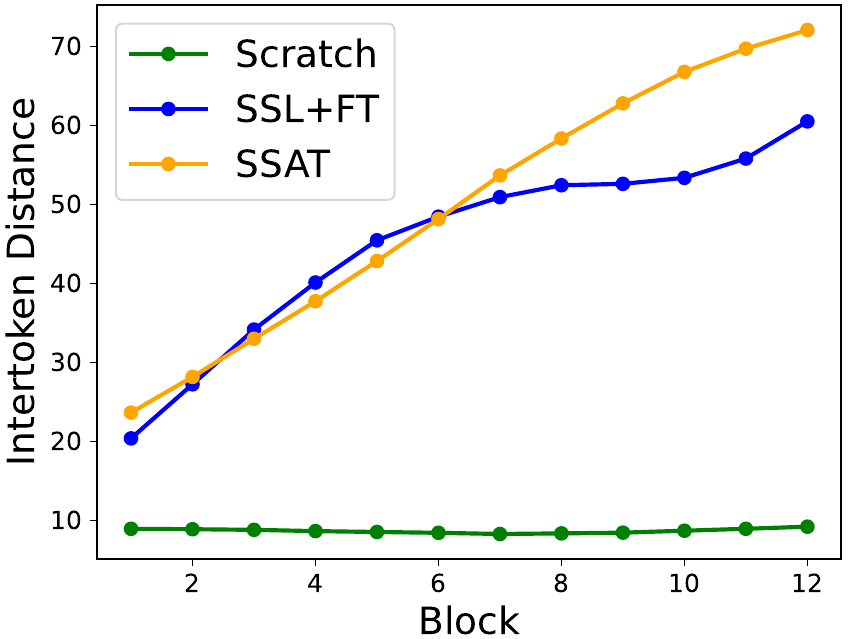}
         \label{ed:flower}
     \end{subfigure}
     \begin{subfigure}[h]{0.2\textwidth}
         \centering
         \includegraphics[width=\textwidth, width=\textwidth]{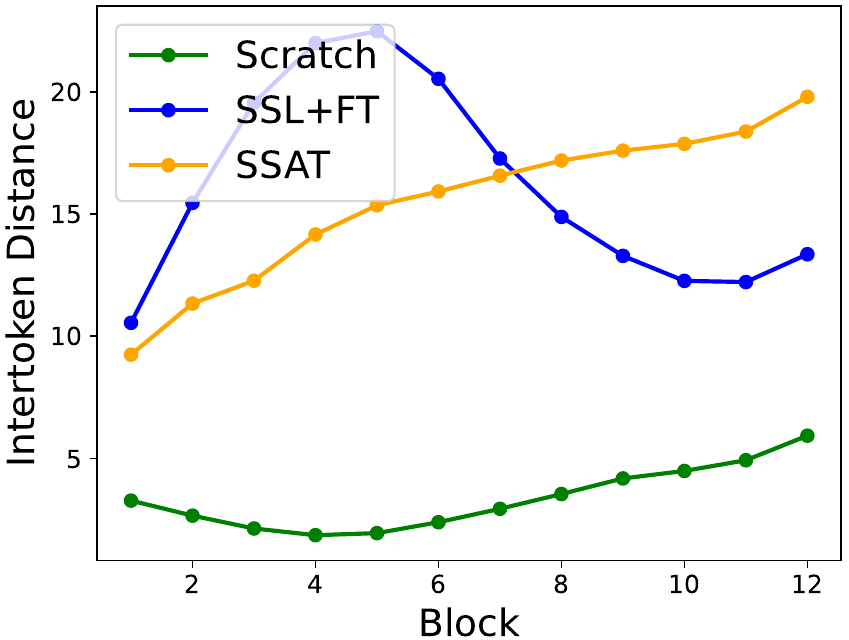}
         \label{ed:c100}
     \end{subfigure}
     \vspace*{-20pt}
    \caption{\textbf{Average Euclidean Inter-token Distance} of ViTs trained from scratch, using SSL+FT and using SSAT, for two different datasets: Oxford Flowers (on the left) and CIFAR-100 (on the right).} \vspace{-0.2in}
    \label{fig:ed}
\end{figure}

\subsection{Diagnosis of features learned by SSAT}
In this section, we differentiate the properties of ViTs learned from scratch, SSL+FT, and our SSAT method. We investigate the learned ViT properties by analyzing their attention weights, token representation, feature transformation, and loss landscape. 
We answer the following key questions:

\noindent \textbf{How are the attention weights distributed?}
The objective of this experiment is to examine the mean attention weights received from other tokens in a sample in the data distribution. As outlined in~\cite{attention_is_all_you_need}, the sum of all values in a column of an $n \times n$ self-attention matrix, where $n$ denotes the number of tokens, represents the aggregated attention associated with a token. Figure~\ref{flower:bin-plots} displays the attention weight distribution across the $n$ tokens for various ViT blocks on both Flower (top row) and CIFAR-100 (bottom row) datasets. The attention weights are uniformly distributed in the first transformer block of the scratch model on both datasets, implying an equal focus on all image regions. However, this distribution changes slightly in the deeper layers. Intriguingly, SSL+FT and SSAT models display sharply peaked attention distributions in the initial and middle transformer layers, but the distributions do not necessarily align with each other. Specifically, in the first transformer block, the attention weight distributions of SSL+FT and SSAT models complement each other, indicating that lower-level features learned by these models are complementary. Moreover, the SSL+FT models exhibit sharp peaks in the final layers, whereas the peaks in SSAT models have a lower magnitude, possibly because the latter model has a better inductive bias. Therefore, although both models are trained on the same set of losses, they use different mechanisms to learn attention weights that differ in the initial layers, and the attention weights learned by the SSAT model are smoother in the final layers, indicating a better inductive bias of the model.

\noindent \textbf{What is the quality of the learned tokens?} 
In this study, we investigated the average distance between tokens within a sample across different transformer blocks. Our analysis involves plotting the average Euclidean distance between tokens in images from the Flower and CIFAR-100 datasets at the output of the transformer layers, as shown in Figure~\ref{fig:ed}. Our results indicate that the scratch model yields a lower inter-token distance than the other models, implying homogeneous token representation. We also observe that SSL+FT models yield higher inter-token distances than SSAT models at the middle transformer layers, but this distance diminishes as we go deeper into the ViTs. Consequently, the SSL+FT models suffer from homogeneous token representation, which adversely affects the ViT training, leading to sub-optimal classification accuracy. In contrast, the inter-token distance of SSAT models increases with ViT depth, indicating that the token representations are discriminative and are semantically rich.
\begin{figure}
     \centering
     \graphicspath{{./images/}}
     \begin{subfigure}[h]{0.2\textwidth}
         \centering
         \includegraphics[width=\textwidth]{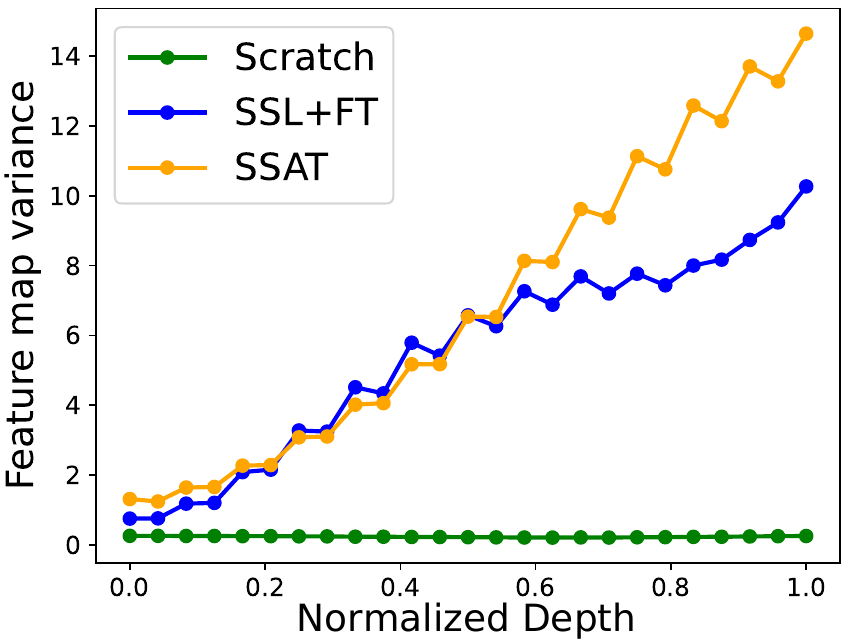}
         \label{variance:flower}
     \end{subfigure}
     \begin{subfigure}[h]{0.2\textwidth}
         \centering
         \includegraphics[width=\textwidth]{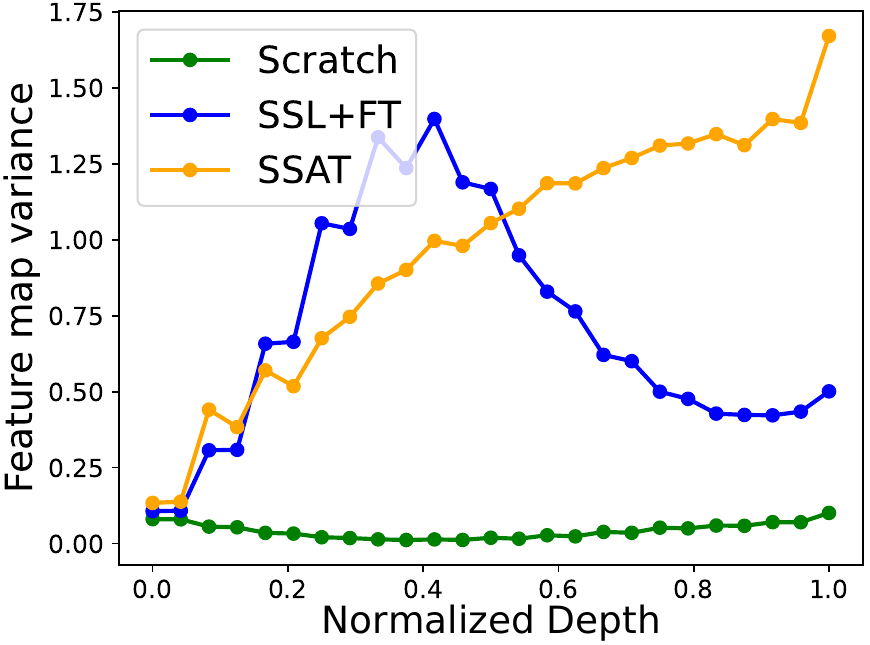}
         \label{variance:c100}
     \end{subfigure}
     \vspace*{-20pt}
    \caption{\textbf{Feature Map Variance} of ViTs trained from scratch, using SSL+FT and using SSAT, for two different datasets: Oxford Flowers (on the left) and CIFAR-100 (on the right).}  \vspace{-0.2in}
    \label{fig:variance}
\end{figure}

\begin{figure*}
     \centering
     \graphicspath{{./images/}}
     \scalebox{0.85}{
     \begin{subfigure}[b]{0.25\textwidth}
         \centering
         \includegraphics[trim={0.35cm 0.35cm 0.35cm 0.35cm},clip,width=\textwidth]{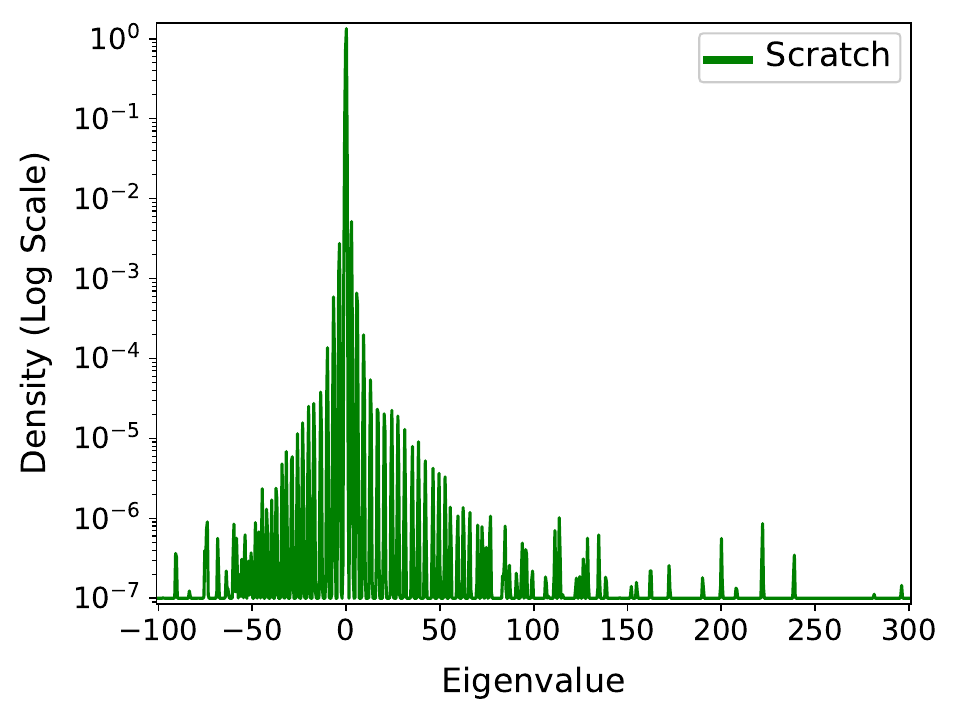}
         \label{fig:scratch}
     \end{subfigure}
     \begin{subfigure}[b]{0.25\textwidth}
         \centering
         \includegraphics[trim={0.35cm 0.35cm 0.35cm 0.35cm},clip,width=\textwidth]{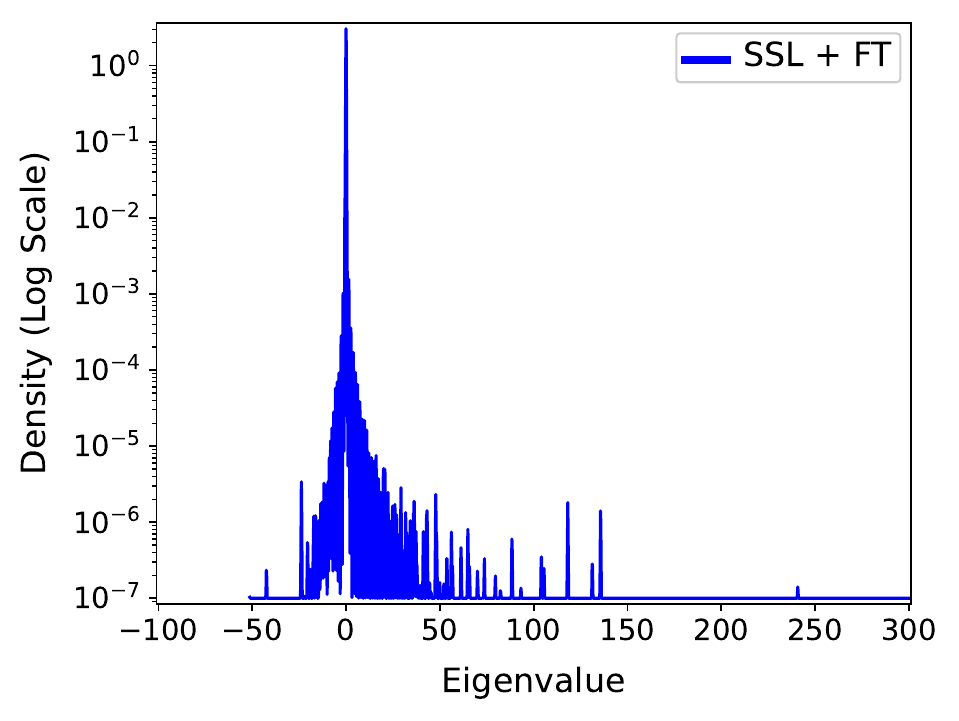}
         \label{fig:hess_ssl}
     \end{subfigure}
     \begin{subfigure}[b]{0.25\textwidth}
         \centering
         \includegraphics[trim={0.35cm 0.35cm 0.35cm 0.35cm},clip, width=\textwidth]{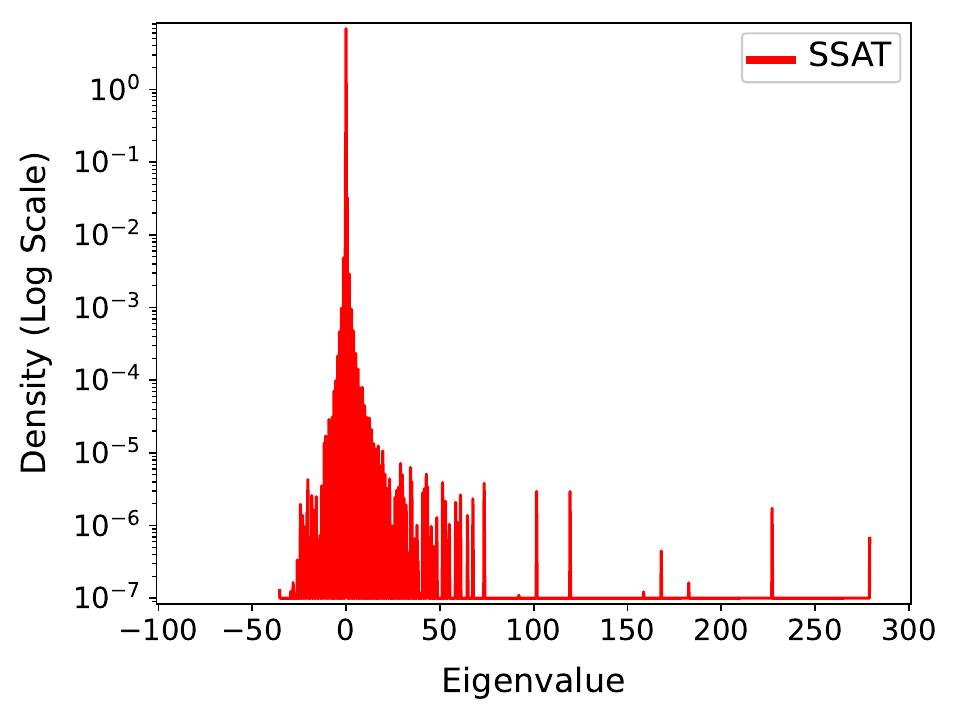}
         \label{fig:hess_ours}
     \end{subfigure}}
    \vspace*{-6mm}
    \caption{\textbf{Hessian max eigenvalues spectra} of ViTs trained from scratch (on the left), SSL + FT (in the middle), and SSAT (on the right).}
    \label{fig:hessian} \vspace{-0.1 in}
\end{figure*}

\noindent \textbf{How are representations transformed?}
The aim of our experiment is to showcase the variation in feature map evolution between ViTs that are trained using different mechanisms. We conducted feature variance measurements across the ViT layers on Flower and CIFAR-100 datasets, and the results are presented in Figure~\ref{fig:variance}. Our analysis confirms the findings of previous studies that the feature variance across ViTs trained from scratch remains constant. However, we observed that the SSL+FT models exhibit an increase in feature variance until a certain layer, after which the rate of an increase either decreases (in Flower dataset) or begins to fall (in CIFAR-100 dataset). Conversely, the feature variance in our SSAT models accumulates with each ViT layer and tends to increase as the depth increases. Consequently, as we go deeper in the SSAT models, the feature map uncertainty decreases, which facilitates optimization through ensembling and stabilizing the transformed feature maps~\cite{pmlr-v162-park22b}.

\noindent \textbf{Why is SSAT better than SSL+FT?}
In this study, we investigated the loss landscapes of ViT models trained using different training mechanisms. We follow~\cite{how_vt} to display the Eigenvalue Spectral Density of Hessian for the different ViT models trained (see Figure~\ref{fig:hessian}). Our results indicate that the scratch ViT model exhibits a wide range of negative Hessian eigenvalues, implying non-convex loss landscapes. Interestingly, the number of negative Hessian eigenvalues is slightly higher in the SSL+FT ViT model than in the scratch model (9622 vs 9667). However, the lower magnitude of some of the negative Hessian eigenvalues in the SSL+FT model makes their qualitative visualization difficult. In contrast, SSAT reduces the number of negative Hessian eigenvalues by 12\% in comparison to the SSL+FT model. This finding suggests that the SSL approach convexifies losses and suppresses negative eigenvalues in the small data regime. Additionally, the SSAT ViT model reduces the average magnitude of negative Hessian eigenvalues by 70\% compared to the SSL+FT models. Therefore, SSAT effectively reduces the magnitude of large Hessian eigenvalues and enhances the ViTs' ability to learn better representations.
\begin{table}
    \centering
    \caption{Comparison of our SSAT to existing state-of-the-art approaches on small datasets. \textsuperscript{\textdagger} indicates that~\cite{smalldata} is replicated with 300 epochs. Results of~\cite{local} is not reported on CIFAR-10.}
    \label{sota}
    \scalebox{0.6}{
    \begin{tabular}{c|c|c|c|c}
    \hline
     \textbf{Method} & \textbf{\# \small{enc. params.}} & \textbf{\small{epochs}} & \textbf{\small{CIFAR-10}} & \textbf{\small{CIFAR-100}}\\
    \hline
     CVT-13+$\mathcal{L}_{drloc}$~\cite{efficient} & \multirow{2}{*}{20M} & \multirow{2}{*}{100} & 90.30 & 74.51 \\
    CVT-13+ SSAT & & & \textbf{95.93} & \textbf{75.16} \\ \hline
     ViT (scratch) & \multirow{4}{*}{2.8M} & \multirow{4}{*}{300} & 93.58 & 73.81 \\ 
     SL-ViT~\cite{small_data_wacv} & & & 94.53 & 76.92 \\
     ViT\textsuperscript{\textdagger}  (SSL+FT)~\cite{smalldata} & & & 94.2 & 76.08 \\
     ViT + SSAT & & & \textbf{95.1} & \textbf{77.8} \\ \hline
     DeiT-Ti+$\mathcal{L}_{guidance}$~\cite{local} & \multirow{2}{*}{6M} & \multirow{2}{*}{300}  & - & 78.15 \\
     DeiT-Ti+$\mathcal{L}_{guidance}$ + SSAT & &  & - & \textbf{79.46} \\
    \hline
    \end{tabular}} \vspace{-0.2 in}
\end{table}

\begin{table*}[t]
\centering
\caption{Cross training evaluation and zero-shot transfer results of DeepFake detection on FaceForensics++ with SSAT. \cite{coccomini2022combining} is trained on both DFDC and FaceForensics++, thus zero-shot transfer results have not been provided.}\vspace{-0.1in}
\scalebox{0.65}{
    \begin{tabular}{c | c  c  c  c | c  c  c  c  c  c} 
     \hline
     \multirow{2}{*}{\textbf{Method}} & \multicolumn{4}{c|}{\textbf{cross-training evaluation}} & \multicolumn{4}{c}{\textbf{zero-shot transfer}}   \\
      & \textbf{Deepfakes} & \textbf{Face2Face} & \textbf{FaceSwap} & \textbf{NeuralTextures} & \textbf{Deepfakes} & \textbf{Face2Face} & \textbf{FaceSwap} & \textbf{NeuralTextures} \\
     \hline
    Scratch & 84.48 & 79.21 & 56.63 & \textbf{82.08} & - & - & - & - \\
     \hline
    Cross-efficient-vit \cite{coccomini2022combining} & 82.67 & 69.89 & 79.93 & 64.87  & - & - & - & - \\
     
    DFDC winner \cite{seferbekov2020dfdc} & 96.43 & 73.93 & 86.07 & 58.57 & 88.57 & 57.50 & 80.36 & 54.64 \\
     \hline
     VideoMAE SSL (0.95) & 82.67 & 64.16 & 58.42 & 63.44 & 86.28 & 49.82 & 69.18 & 51.97 \\ 
     
     VideoMAE SSL (0.75) & 78.34 & 65.59 & 57.35 & 61.65  & 82.67 & 48.39 & 65.23 & 51.97 \\
     \hline
     \rowcolor{Gray} \textbf{VideoMAE (0.95) + SSAT} & 92.42 & 79.21 & 89.61 & 81.36  & \textbf{92.42} & \textbf{61.65} & \textbf{92.83} & \textbf{62.37} \\ 
     
      \rowcolor{Gray} \textbf{VideoMAE (0.75) + SSAT} & \textbf{96.75} & \textbf{80.65} & \textbf{91.40} & 72.76 & 87.73 & 60.57 & 88.17 & 61.65 \\ 
     
     \hline
    \end{tabular}}%
    \vspace{-0.2in}
    \label{tab:ff}
\end{table*}
\subsection{Comparison with the state-of-the-art}
Table~\ref{sota} presents a comparison of SSAT with state-of-the-art (SOTA) methods. To ensure a fair evaluation, we implemented SSAT with the ViT encoder used in the respective methods. We find that MAE as SSAT outperforms Drloc~\cite{efficient} which takes predicting relative distance between patches as SSAT. This shows that the choice of SSAT plays a crucial role in the effective training of ViTs. Moreover, we find that SSAT outperforms SL-ViT~\cite{small_data_wacv} and~\cite{smalldata} when trained for an equal number of epochs. This indicates that SSAT, without any architectural modifications, can surpass SOTA methods through its joint training strategy. Additionally, we trained a ViT with SSAT and feature-level distillation from a light-weight CNN as described in~\cite{local}. The improvement over the baseline~\cite{local}, which involves training ViT with feature-level distillation only, demonstrates the complementary nature of the representations learned by ViT when trained with SSAT. 

\begin{figure}[h!]
\centering
\scalebox{0.8}{
  \includegraphics[width=1\linewidth]{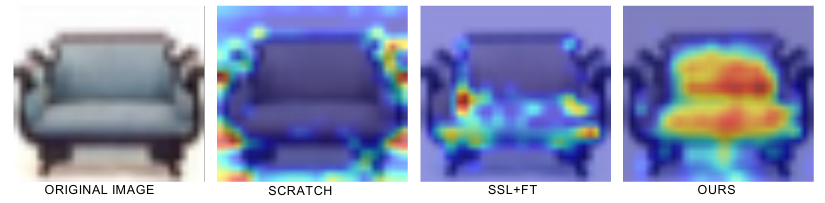}}\vspace{-0.1in}
   \caption{\textbf{GradCAM visualizations} of our SSAT model and the representative baselines.}
   \label{gradcam}
   \vspace{-5mm} 
\end{figure}

In Figure~\ref{gradcam}, we present the Grad-CAM visualizations~\cite{grad_cam}. SSL+FT (3rd col) focuses on few specific  pixelwise regions, while our method (4th col) focuses on areas corresponding to the entire primary object. We also provide the attention visualization of ViTs trained using different strategies in Appendix~\ref{vis_appendix} (see Figure~\ref{attention}).
\vspace{-0.1in}

\subsection{Performance of SSAT in video domain}
We have also assessed the efficacy of the SSAT within the video domain for the task of deepfake detection. In this experiment, the model's generalization capabilities for deepfake detection is validated, as presented in Table~\ref{tab:ff}. For video encoding, we employ ViT as in~\cite{videomae}. Our VideoMAE + SSAT model is a direct extension of the MAE+SSAT model designed for image data; the only modification lies in the choice of encoder. The primary task involves binary classification to distinguish between real and manipulated videos. Notably, we experimented with two masking ratios, 0.75 and 0.95, during the training of VideoMAE + SSAT.

To assess model generalizability, we conducted cross-manipulation training based on the FaceForensics++ dataset~\cite{das2021demystifying}. We trained the model using videos generated by all possible combinations of three manipulation techniques (Deepfakes, Face2Face, FaceSwap, and NeuralTextures) plus original videos, and then evaluated its performance on the videos generated by the remaining technique. This approach simulates real-world scenarios where multiple manipulation techniques might be encountered post-training. All models, except the scratch model, are pretrained on the DFDC dataset~\cite{seferbekov2020dfdc} before being evaluated on the FaceForensics++ dataset. To enable a fair comparison with VideoMAE + SSAT, the baseline VideoMAE SSL models are first pretrained and fine-tuned on DFDC, 
and are subsequently employed for deepfake classification task. 
The evaluation involved both (1) cross-dataset fine-tuning on FaceForensics++ and (2) zero-shot transfer assessment where pre-trained models are evaluated on FaceForensics++ without additional training.

Our findings, as detailed in Table~\ref{tab:ff}, reveal that the VideoMAE+SSAT models demonstrate a superior generalized capability than the other baselines to distinguish between real and manipulated videos. Note that the scratch model outperforms all models on detecting videos generated using NeuralTextures without any pretraining but it is not suitable for zeros-shot transfer. Interestingly, the VideoMAE models exhibit complementary behavior when subjected to different masking ratios, which warrants a future investigation. More details including implementation and training details of these experiments can be found in Appendix~\ref{deepfake_appendix}.

\section{Conclusion}
The main focus of this paper was on the use of self-supervised learning (SSL) to effectively train ViTs on domains with limited data. We demonstrate that by jointly optimizing the primary task of a ViT encoder with SSL as an auxiliary task, we can achieve discriminative representations for the primary task. This simple and easy-to-implement method called SSAT outperforms the traditional approach of sequentially training with SSL followed by fine-tuning on the same data. Our joint training framework learns features that are different from those learned by the dissociated framework, even when using the same losses. These results highlight the potential of SSAT as an effective training strategy with a lower carbon footprint. We anticipate that SSAT will become a standard norm for training vision transformers on small datasets. 

\section*{Acknowledgments}
We thank the members of the Charlotte Machine Learning Lab at UNC Charlotte for valuable discussion. 
This work is supported by the National Science Foundation (IIS-2245652). 

\newpage
{\small
\bibliographystyle{ieee_fullname}
\bibliography{egbib}

\begin{thebibliography}{10}\itemsep=-1pt

\bibitem{arnab2021vivit}
Anurag Arnab, Mostafa Dehghani, Georg Heigold, Chen Sun, Mario Lu{\v{c}}i{\'c},
  and Cordelia Schmid.
\newblock {VIVIT: A video vision transformer}.
\newblock {\em arXiv preprint arXiv:2103.15691}, 2021.

\bibitem{timesformer}
Gedas Bertasius, Heng Wang, and Lorenzo Torresani.
\newblock Is space-time attention all you need for video understanding?
\newblock In {\em Int. Conf. on Mach. Learn.}, July 2021.

\bibitem{buslaev2020albumentations}
Alexander Buslaev, Vladimir~I Iglovikov, Eugene Khvedchenya, Alex Parinov,
  Mikhail Druzhinin, and Alexandr~A Kalinin.
\newblock Albumentations: fast and flexible image augmentations.
\newblock {\em Information}, 11(2):125, 2020.

\bibitem{dino}
Mathilde Caron, Hugo Touvron, Ishan Misra, Herv{\'e} J{\'e}gou, Julien Mairal,
  Piotr Bojanowski, and Armand Joulin.
\newblock Emerging properties in self-supervised vision transformers.
\newblock In {\em Proceedings of the IEEE/CVF International Conference on
  Computer Vision}, pages 9650--9660, 2021.

\bibitem{hipt}
Richard~J Chen, Chengkuan Chen, Yicong Li, Tiffany~Y Chen, Andrew~D Trister,
  Rahul~G Krishnan, and Faisal Mahmood.
\newblock Scaling vision transformers to gigapixel images via hierarchical
  self-supervised learning.
\newblock In {\em Proceedings of the IEEE/CVF Conference on Computer Vision and
  Pattern Recognition}, pages 16144--16155, 2022.

\bibitem{chen2022self}
Richard~J Chen and Rahul~G Krishnan.
\newblock Self-supervised vision transformers learn visual concepts in
  histopathology.
\newblock {\em arXiv preprint arXiv:2203.00585}, 2022.

\bibitem{simclr}
Ting Chen, Simon Kornblith, Mohammad Norouzi, and Geoffrey Hinton.
\newblock A simple framework for contrastive learning of visual
  representations.
\newblock In {\em International conference on machine learning}, pages
  1597--1607. PMLR, 2020.

\bibitem{cheng2021per}
Bowen Cheng, Alexander~G Schwing, and Alexander Kirillov.
\newblock Per-pixel classification is not all you need for semantic
  segmentation.
\newblock {\em arXiv preprint arXiv:2107.06278}, 2021.

\bibitem{coccomini2022combining}
Davide~Alessandro Coccomini, Nicola Messina, Claudio Gennaro, and Fabrizio
  Falchi.
\newblock Combining efficientnet and vision transformers for video deepfake
  detection.
\newblock In {\em International conference on image analysis and processing},
  pages 219--229. Springer, 2022.

\bibitem{dai2022mstct}
Rui Dai, Srijan Das, Kumara Kahatapitiya, Michael Ryoo, and Francois Bremond.
\newblock {MS-TCT: Multi-Scale Temporal ConvTransformer for Action Detection}.
\newblock In {\em CVPR}, 2022.

\bibitem{das2021demystifying}
Abhijit Das, Srijan Das, and Antitza Dantcheva.
\newblock Demystifying attention mechanisms for deepfake detection.
\newblock In {\em 2021 16th IEEE International Conference on Automatic Face and
  Gesture Recognition (FG 2021)}, pages 1--7. IEEE, 2021.

\bibitem{imagenet_cvpr09}
Jia Deng, Wei Dong, Richard Socher, Li-Jia Li, Kai Li, and Li Fei-Fei.
\newblock {ImageNet: A Large-Scale Hierarchical Image Database}.
\newblock In {\em CVPR09}, 2009.

\bibitem{dosovitskiy2020vit}
Alexey Dosovitskiy, Lucas Beyer, Alexander Kolesnikov, Dirk Weissenborn,
  Xiaohua Zhai, Thomas Unterthiner, Mostafa Dehghani, Matthias Minderer, Georg
  Heigold, Sylvain Gelly, Jakob Uszkoreit, and Neil Houlsby.
\newblock An image is worth 16x16 words: Transformers for image recognition at
  scale.
\newblock {\em ICLR}, 2021.

\bibitem{mvit1}
Haoqi Fan, Bo Xiong, Karttikeya Mangalam, Yanghao Li, Zhicheng Yan, Jitendra
  Malik, and Christoph Feichtenhofer.
\newblock Multiscale vision transformers.
\newblock In {\em ICCV}, 2021.

\bibitem{smalldata}
Hanan Gani, Muzammal Naseer, and Mohammad Yaqub.
\newblock How to train vision transformer on small-scale datasets?
\newblock {\em arXiv preprint arXiv:2210.07240}, 2022.

\bibitem{convmae}
Peng Gao, Teli Ma, Hongsheng Li, Jifeng Dai, and Yu Qiao.
\newblock Convmae: Masked convolution meets masked autoencoders.
\newblock {\em arXiv preprint arXiv:2205.03892}, 2022.

\bibitem{byol}
Jean-Bastien Grill, Florian Strub, Florent Altch{\'e}, Corentin Tallec, Pierre
  Richemond, Elena Buchatskaya, Carl Doersch, Bernardo Avila~Pires, Zhaohan
  Guo, Mohammad Gheshlaghi~Azar, et~al.
\newblock Bootstrap your own latent-a new approach to self-supervised learning.
\newblock {\em Advances in neural information processing systems},
  33:21271--21284, 2020.

\bibitem{tnt}
Kai Han, An Xiao, Enhua Wu, Jianyuan Guo, Chunjing Xu, and Yunhe Wang.
\newblock Transformer in transformer, 2021.

\bibitem{MaskedAutoencoders2021}
Kaiming He, Xinlei Chen, Saining Xie, Yanghao Li, Piotr Doll{\'a}r, and Ross
  Girshick.
\newblock Masked autoencoders are scalable vision learners.
\newblock {\em arXiv:2111.06377}, 2021.

\bibitem{mae}
Kaiming He, Xinlei Chen, Saining Xie, Yanghao Li, Piotr Doll{\'a}r, and Ross
  Girshick.
\newblock Masked autoencoders are scalable vision learners.
\newblock In {\em Proceedings of the IEEE/CVF Conference on Computer Vision and
  Pattern Recognition}, pages 16000--16009, 2022.

\bibitem{moco}
Kaiming He, Haoqi Fan, Yuxin Wu, Saining Xie, and Ross Girshick.
\newblock Momentum contrast for unsupervised visual representation learning.
\newblock In {\em Proceedings of the IEEE/CVF conference on computer vision and
  pattern recognition}, pages 9729--9738, 2020.

\bibitem{resnet-50}
Kaiming He, Xiangyu Zhang, Shaoqing Ren, and Jian Sun.
\newblock Deep residual learning for image recognition.
\newblock In {\em 2016 IEEE Conference on Computer Vision and Pattern
  Recognition (CVPR)}, pages 770--778, 2016.

\bibitem{heo2021pit}
Byeongho Heo, Sangdoo Yun, Dongyoon Han, Sanghyuk Chun, Junsuk Choe, and
  Seong~Joon Oh.
\newblock Rethinking spatial dimensions of vision transformers.
\newblock In {\em International Conference on Computer Vision (ICCV)}, 2021.

\bibitem{cdnet_med}
Saarthak Kapse, Srijan Das, and Prateek Prasanna.
\newblock Cd-net: Histopathology representation learning using pyramidal
  context-detail network, 2022.

\bibitem{cifar}
Alex Krizhevsky, Geoffrey Hinton, et~al.
\newblock Learning multiple layers of features from tiny images.
\newblock 2009.

\bibitem{co_emmission}
Alexandre Lacoste, Alexandra Luccioni, Victor Schmidt, and Thomas Dandres.
\newblock Quantifying the carbon emissions of machine learning.
\newblock {\em arXiv preprint arXiv:1910.09700}, 2019.

\bibitem{laskin2020curl}
Michael Laskin, Aravind Srinivas, and Pieter Abbeel.
\newblock Curl: Contrastive unsupervised representations for reinforcement
  learning.
\newblock In {\em Int. Conf. on Mach. Learn.}, pages 5639--5650. PMLR, 2020.

\bibitem{small_data_wacv}
Seung~Hoon Lee, Seunghyun Lee, and Byung~Cheol Song.
\newblock Vision transformer for small-size datasets.
\newblock {\em CoRR}, abs/2112.13492, 2021.

\bibitem{local}
Kehan Li, Runyi Yu, Zhennan Wang, Li Yuan, Guoli Song, and Jie Chen.
\newblock Locality guidance for improving vision transformers on tiny datasets.
\newblock In {\em Computer Vision--ECCV 2022: 17th European Conference, Tel
  Aviv, Israel, October 23--27, 2022, Proceedings, Part XXIV}, pages 110--127.
  Springer, 2022.

\bibitem{lidoes}
Xiang Li, Jinghuan Shang, Srijan Das, and Michael~S Ryoo.
\newblock Does self-supervised learning really improve reinforcement learning
  from pixels?
\newblock In {\em Advances in Neural Information Processing Systems}, 2022.

\bibitem{uniform_masking}
Xiang Li, Wenhai Wang, Lingfeng Yang, and Jian Yang.
\newblock Uniform masking: Enabling mae pre-training for pyramid-based vision
  transformers with locality.
\newblock {\em arXiv:2205.10063}, 2022.

\bibitem{mvit2}
Yanghao Li, Chao-Yuan Wu, Haoqi Fan, Karttikeya Mangalam, Bo Xiong, Jitendra
  Malik, and Christoph Feichtenhofer.
\newblock Mvitv2: Improved multiscale vision transformers for classification
  and detection.
\newblock In {\em CVPR}, 2022.

\bibitem{efficient}
Yahui Liu, Enver Sangineto, Wei Bi, Nicu Sebe, Bruno Lepri, and Marco De~Nadai.
\newblock Efficient training of visual transformers with small datasets.
\newblock In {\em Conference on Neural Information Processing Systems
  (NeurIPS)}, 2021.

\bibitem{swinver1}
Ze Liu, Yutong Lin, Yue Cao, Han Hu, Yixuan Wei, Zheng Zhang, Stephen Lin, and
  Baining Guo.
\newblock Swin transformer: Hierarchical vision transformer using shifted
  windows.
\newblock In {\em CVPR}, pages 10012--10022, 2021.

\bibitem{liu2021video}
Ze Liu, Jia Ning, Yue Cao, Yixuan Wei, Zheng Zhang, Stephen Lin, and Han Hu.
\newblock Video swin transformer.
\newblock {\em arXiv preprint arXiv:2106.13230}, 2021.

\bibitem{svhn}
Yuval Netzer, Tao Wang, Adam Coates, Alessandro Bissacco, Bo Wu, and Andrew~Y.
  Ng.
\newblock Reading digits in natural images with unsupervised feature learning.
\newblock In {\em NIPS Workshop on Deep Learning and Unsupervised Feature
  Learning 2011}, 2011.

\bibitem{flowers}
Maria-Elena Nilsback and Andrew Zisserman.
\newblock Automated flower classification over a large number of classes.
\newblock In {\em 2008 Sixth Indian Conference on Computer Vision, Graphics \&
  Image Processing}, pages 722--729. IEEE, 2008.

\bibitem{pmlr-v162-park22b}
Namuk Park and Songkuk Kim.
\newblock Blurs behave like ensembles: Spatial smoothings to improve accuracy,
  uncertainty, and robustness.
\newblock In Kamalika Chaudhuri, Stefanie Jegelka, Le Song, Csaba Szepesvari,
  Gang Niu, and Sivan Sabato, editors, {\em Proceedings of the 39th
  International Conference on Machine Learning}, volume 162 of {\em Proceedings
  of Machine Learning Research}, pages 17390--17419. PMLR, 17--23 Jul 2022.

\bibitem{how_vt}
Namuk Park and Songkuk Kim.
\newblock How do vision transformers work?
\newblock {\em arXiv preprint arXiv:2202.06709}, 2022.

\bibitem{ssl_learn}
Namuk Park, Wonjae Kim, Byeongho Heo, Taekyung Kim, and Sangdoo Yun.
\newblock What do self-supervised vision transformers learn?
\newblock In {\em International Conference on Learning Representations}.

\bibitem{domainnet}
Xingchao Peng, Qinxun Bai, Xide Xia, Zijun Huang, Kate Saenko, and Bo Wang.
\newblock Moment matching for multi-source domain adaptation.
\newblock In {\em Proceedings of the IEEE/CVF international conference on
  computer vision}, pages 1406--1415, 2019.

\bibitem{evolving_losses}
AJ Piergiovanni, Anelia Angelova, and Michael~S. Ryoo.
\newblock Evolving losses for unsupervised video representation learning.
\newblock In {\em IEEE/CVF Conference on Computer Vision and Pattern
  Recognition (CVPR)}, June 2020.

\bibitem{ranasinghe2022self}
Kanchana Ranasinghe, Muzammal Naseer, Salman Khan, Fahad~Shahbaz Khan, and
  Michael~S Ryoo.
\newblock Self-supervised video transformer.
\newblock In {\em Proceedings of the IEEE/CVF Conference on Computer Vision and
  Pattern Recognition}, pages 2874--2884, 2022.

\bibitem{roessler2019faceforensicspp}
Andreas R\"ossler, Davide Cozzolino, Luisa Verdoliva, Christian Riess, Justus
  Thies, and Matthias Nie{\ss}ner.
\newblock Face{F}orensics++: Learning to detect manipulated facial images.
\newblock In {\em International Conference on Computer Vision (ICCV)}, 2019.

\bibitem{ryoo2021tokenlearner}
Michael Ryoo, AJ Piergiovanni, Anurag Arnab, Mostafa Dehghani, and Anelia
  Angelova.
\newblock {TokenLearner: Adaptive Space-Time Tokenization for Videos}.
\newblock {\em Advances in Neural Information Processing Systems}, 34, 2021.

\bibitem{seferbekov2020dfdc}
Selim Seferbekov.
\newblock Dfdc 1st place solution, 2020.

\bibitem{grad_cam}
Ramprasaath~R. {Selvaraju}, Michael {Cogswell}, Abhishek {Das}, Ramakrishna
  {Vedantam}, Devi {Parikh}, and Dhruv {Batra}.
\newblock Grad-cam: Visual explanations from deep networks via gradient-based
  localization.
\newblock In {\em 2017 IEEE International Conference on Computer Vision
  (ICCV)}, pages 618--626, 2017.

\bibitem{3dtrl}
Jinghuan Shang, Srijan Das, and Michael~S Ryoo.
\newblock Learning viewpoint-agnostic visual representations by recovering
  tokens in 3d space.
\newblock In {\em Advances in Neural Information Processing Systems}, 2022.

\bibitem{scorenet}
Thomas Stegm{\"u}ller, Antoine Spahr, Behzad Bozorgtabar, and Jean-Philippe
  Thiran.
\newblock Scorenet: Learning non-uniform attention and augmentation for
  transformer-based histopathological image classification.
\newblock {\em arXiv preprint arXiv:2202.07570}, 2022.

\bibitem{jft_300m}
Chen Sun, Abhinav Shrivastava, Saurabh Singh, and Abhinav Gupta.
\newblock Revisiting unreasonable effectiveness of data in deep learning era.
\newblock In {\em ICCV}, pages 843--852. IEEE Computer Society, 2017.

\bibitem{videomae}
Zhan Tong, Yibing Song, Jue Wang, and Limin Wang.
\newblock Videomae: Masked autoencoders are data-efficient learners for
  self-supervised video pre-training.
\newblock {\em arXiv preprint arXiv:2203.12602}, 2022.

\bibitem{deit}
Hugo Touvron, Matthieu Cord, Matthijs Douze, Francisco Massa, Alexandre
  Sablayrolles, and Herve Jegou.
\newblock Training data-efficient image transformers: Distillation through
  attention.
\newblock In {\em Int. Conf. on Mach. Learn.}, volume 139, pages 10347--10357,
  July 2021.

\bibitem{attention_is_all_you_need}
Ashish Vaswani, Noam Shazeer, Niki Parmar, Jakob Uszkoreit, Llion Jones,
  Aidan~N. Gomez, Lukasz Kaiser, and Illia Polosukhin.
\newblock Attention is all you need.
\newblock In {\em NIPS}, 2017.

\bibitem{wang2021pvtv2}
Wenhai Wang, Enze Xie, Xiang Li, Deng-Ping Fan, Kaitao Song, Ding Liang, Tong
  Lu, Ping Luo, and Ling Shao.
\newblock {PVTv2: Improved baselines with pyramid vision transformer}.
\newblock {\em arXiv preprint arXiv:2106.13797}, 2021.

\bibitem{wang2021pyramid}
Wenhai Wang, Enze Xie, Xiang Li, Deng-Ping Fan, Kaitao Song, Ding Liang, Tong
  Lu, Ping Luo, and Ling Shao.
\newblock Pyramid vision transformer: A versatile backbone for dense prediction
  without convolutions.
\newblock {\em arXiv preprint arXiv:2102.12122}, 2021.

\bibitem{transpath}
Xiyue Wang, Sen Yang, Jun Zhang, Minghui Wang, Jing Zhang, Junzhou Huang, Wei
  Yang, and Xiao Han.
\newblock Transpath: Transformer-based self-supervised learning for
  histopathological image classification.
\newblock In {\em International Conference on Medical Image Computing and
  Computer-Assisted Intervention}, pages 186--195. Springer, 2021.

\bibitem{cvt}
Haiping Wu, Bin Xiao, Noel Codella, Mengchen Liu, Xiyang Dai, Lu Yuan, and Lei
  Zhang.
\newblock Cvt: Introducing convolutions to vision transformers.
\newblock In {\em Proceedings of the IEEE/CVF International Conference on
  Computer Vision}, pages 22--31, 2021.

\bibitem{xie2021segformer}
Enze Xie, Wenhai Wang, Zhiding Yu, Anima Anandkumar, Jose~M Alvarez, and Ping
  Luo.
\newblock Segformer: Simple and efficient design for semantic segmentation with
  transformers.
\newblock {\em arXiv preprint arXiv:2105.15203}, 2021.

\bibitem{simmim}
Zhenda Xie, Zheng Zhang, Yue Cao, Yutong Lin, Jianmin Bao, Zhuliang Yao, Qi
  Dai, and Han Hu.
\newblock Simmim: A simple framework for masked image modeling.
\newblock In {\em Proceedings of the IEEE/CVF Conference on Computer Vision and
  Pattern Recognition}, pages 9653--9663, 2022.

\bibitem{medmnist}
Jiancheng Yang, Rui Shi, Donglai Wei, Zequan Liu, Lin Zhao, Bilian Ke,
  Hanspeter Pfister, and Bingbing Ni.
\newblock Medmnist v2-a large-scale lightweight benchmark for 2d and 3d
  biomedical image classification.
\newblock {\em Scientific Data}, 10(1):41, 2023.

\bibitem{t2t}
Li Yuan, Yunpeng Chen, Tao Wang, Weihao Yu, Yujun Shi, Zi-Hang Jiang,
  Francis~EH Tay, Jiashi Feng, and Shuicheng Yan.
\newblock Tokens-to-token vit: Training vision transformers from scratch on
  imagenet.
\newblock In {\em Proceedings of the IEEE/CVF international conference on
  computer vision}, pages 558--567, 2021.

\bibitem{zhou2021deepvit}
Daquan Zhou, Bingyi Kang, Xiaojie Jin, Linjie Yang, Xiaochen Lian, Zihang
  Jiang, Qibin Hou, and Jiashi Feng.
\newblock Deepvit: Towards deeper vision transformer.
\newblock {\em arXiv preprint arXiv:2103.11886}, 2021.

\bibitem{chaoyang}
Chuang Zhu, Wenkai Chen, Ting Peng, Ying Wang, and Mulan Jin.
\newblock Hard sample aware noise robust learning for histopathology image
  classification.
\newblock {\em IEEE Transactions on Medical Imaging}, 41(4):881--894, 2021.

\bibitem{zhu2020deformable}
Xizhou Zhu, Weijie Su, Lewei Lu, Bin Li, Xiaogang Wang, and Jifeng Dai.
\newblock Deformable detr: Deformable transformers for end-to-end object
  detection.
\newblock {\em arXiv preprint arXiv:2010.04159}, 2020.

\end{thebibliography}
}

\clearpage

\appendix
\noindent \textbf{\huge Appendix}

\section{Dataset Description}
This paper presents a comprehensive evaluation of our ViT models on 10 different image datasets, comprising prominent computer vision benchmarks such as ImageNet-1K~\cite{imagenet_cvpr09} (IN-1K), CIFAR-10 and CIFAR-100~\cite{cifar}, Oxford Flowers102~\cite{flowers}, and SVHN~\cite{svhn}. In addition, we include three datasets namely ClipArt, Infograph, and Sketch from DomainNet~\cite{domainnet}, a widely adopted benchmark for domain adaptation tasks. Moreover, we explore the performance of our approach on two medical image domain datasets: Chaoyang~\cite{chaoyang} and PneumoniaMNIST~\cite{medmnist}. The dataset size, sample resolution, and the number of classes are further elaborated in Table~\ref{datasets}. Note that the accuracies reported for CIFAR in Figure 1 of the main paper is an average of the classification accuracy of CIFAR-10 and CIFAR-100.

\begin{table}[!h]
    \centering
    \caption{Details of image classification datasets (sample size, resolution, and number of classes) evaluated in our experiments. }
    \resizebox{0.47\textwidth}{!}{
    \begin{tabular}{lcccc}
    \toprule
    \textbf{Dataset} & \textbf{Train Size} & \textbf{Test Size} &  \textbf{Dimensions} & \textbf{\# Classes}\\
    \midrule
    CIFAR-10 & 50,000 & 10,000 & 32$\times$32 & 10\\
    CIFAR-100 & 50,000 & 10,000 & 32$\times$32 & 100\\
    Flowers102 & 2,040 & 6,149 & 224$\times$224 & 102 \\
    SVHN & 73,257 & 26,032 & 32$\times$32 & 10 \\
    \midrule
    ImageNet-1K & 1,281,167 & 100,000 & 224$\times$224 & 1000\\
    \midrule
    ClipArt & 33,525 & 14,604\\
    Infograph & 36,023 & 15,582 & 224$\times$224 & 345\\
    Sketch & 50,416 & 21,850\\
    \midrule
    Chaoyang & 4,021 & 2,139 & 512$\times$512 & 4\\
    PMNIST & 5,232 & 624 & 28$\times$28 & 2 \\
    \bottomrule
    \end{tabular}}
    \label{datasets}
\end{table}
\begin{table}[!h]
    \caption{Ablation of decoder depth.} 
    \centering 
    \begin{tabular}{c|ccccc} %
    \toprule
    \multirow{2}{*}{
    \parbox[c]{.2\linewidth}{\centering \textbf{Decoder Depth}}} & \multicolumn{2}{c}{\textbf{Accuracy}} \\ 
    \cmidrule{2-3}
     & {\centering \textbf{CIFAR-10}} & {\textbf{CIFAR-100}}\\
    \midrule
    1 & 91.59 & 68.41\\
    2 & \textbf{91.65} & \textbf{69.64}\\
    4 & 90.88 & 67.46 \\
    8 & 90.59 & 67.78\\
    12 & 91.08 & 66.94\\
    \bottomrule
    \end{tabular}
    \label{ablation:decoder_depth}
\end{table}
\begin{table}[!h]
    \caption{Ablation of decoder embedding dimension.} 
    \centering 
    \begin{tabular}{c|ccccc} %
    \toprule
    \multirow{2}{*}{
    \parbox[c]{0.25\linewidth}{\centering \textbf{Decoder Dimension}}} & \multicolumn{2}{c}{\textbf{Accuracy}} \\ 
    \cmidrule{2-3}
     & {\centering \textbf{CIFAR-10}} & {\textbf{CIFAR-100}}\\
    \midrule
    64 & 89.20 & 67.11 \\
    128 & \textbf{91.65} & \textbf{69.64}\\
    256 & 91.64 & 66.98\\
    512 & 90.53 & 66.19\\
    \bottomrule
    \end{tabular}
    \label{ablation:decoder_dimension}
\end{table}
\begin{table}[!h]
    \caption{Ablation of Decoder Heads.} 
    \centering 
    \begin{tabular}{cccccc} %
    \toprule
    \multirow{2}{*}{
    \parbox[c]{0.2\linewidth}{\centering Decoder Heads}} & \multicolumn{2}{c}{Accuracy} \\ 
    \cmidrule{2-3}
     & {\centering CIFAR-10} & {CIFAR-100}\\
    \midrule
    1 & 91.54 & 69.44\\
    2 & 92.44 & 69.28\\
    4 & \textbf{92.49} & 68.59\\
    8 & 92.09 &  69.52\\
    16 & 91.65 & \textbf{69.64}\\
    \bottomrule
    \end{tabular}
    \label{ablation:decoder_heads}
\end{table}

\begin{table}
\caption{Statistics of video datasets generated by different manipulating techniques available in Faceforensics++}
\scalebox{0.65}{
\begin{tabular}{c | c  c  c  c  c | c } 
 \toprule
 \textbf{Split} & \textbf{DeepFake} & \textbf{Face2Face} & \textbf{FaceSwap} & \textbf{NeuralTextures} & \textbf{Original} & \textbf{Total} \\ 
 \midrule
\textbf{Train}         & 720               & 720                & 720               & 720                     & 720               & 3600 \\ 
\textbf{Val}           & 140               & 140                & 140               & 140                     & 140               & 700 \\
\textbf{Test}          & 140               & 140                & 140               & 140                     & 140               & 700 \\
 \midrule
 \textbf{Total}   & 1000              & 1000               & 1000              & 1000                    & 1000              & 5000 \\
 \bottomrule
\end{tabular}%
}
\label{tab:stat_ff++}
\end{table}

\begin{table*}[h]
    \centering
    \caption{Our ViT training settings across different datasets.}
    \resizebox{0.9\textwidth}{!}{
    \begin{tabular}{l|cl}
    \toprule
    
    \multirow{5}{*}{Input Size} & PMNIST  & {\Big|} \hspace{0.1215cm} 28$\times$28 \\
    \cmidrule(lr){2-3}
    & CIFAR10, CIFAR100, SVHN  & {\Big|} \hspace{0.1215cm} 32$\times$32 \\
    \cmidrule(lr){2-3}
    & \multirow{2}{*}{\makecell{Flowers, ImageNet-1K\\ClipArt, Infograph, Sketch}} & \multirow{2}{*}{\Big|} \multirow{2}{*}{224$\times$224} \\ \\
    \cmidrule(lr){2-3}
    & Chaoyang & {\Big|} \hspace{0.1215cm} 512$\times$512 \\ \midrule
    
    \multirow{4}{*}{Patch Size} & \multirow{2}{*}{\makecell{PMNIST, CIFAR10, \\ 
    CIFAR100, SVHN}} & \multirow{2}{*}{\Big|} \multirow{2}{*}{2$\times$2} \\ \\
    \cmidrule(lr){2-3}
    & \multirow{2}{*}{\makecell{Flowers, ImageNet-1K\\ClipArt, Infograph, Sketch}} & \multirow{2}{*}{\Big|} \multirow{2}{*}{16$\times$16} \\ \\
    \cmidrule(lr){2-3}
    & Chaoyang & {\Big|} \hspace{0.1215cm} 32$\times$32 \\ \midrule
    
    Batch Size & 64 \\ \midrule
    Optimizer & \multicolumn{1}{c}{AdamW}\\
    Optimizer Epsilon &  \multicolumn{1}{c}{1e-08}\\
    Momentum & \multicolumn{1}{c}{$\beta_{1}=0.9$,
    \:$\beta_{2}=0.999$}\\
    layer-wise lr decay & \multicolumn{1}{c}{0.75}\\
    Weight Decay & \multicolumn{1}{c}{0.05}\\
    Gradient Clip & \multicolumn{1}{c}{None}\\
    \midrule
    Learning Rate Schedule & \multicolumn{1}{c}{Cosine}\\
    Learning Rate & \multicolumn{1}{c}{1e-03}\\
    Warmup LR  & \multicolumn{1}{c}{1e-06}\\
    Min LR & \multicolumn{1}{c}{1e-6}\\
    Epochs & \multicolumn{1}{c}{100}\\
    Warmup Epochs  & \multicolumn{1}{c}{5}\\
    Decay Rate  & \multicolumn{1}{c}{0.05}\\
    Drop Path & \multicolumn{1}{c}{0.01}\\
    Lambda ($\lambda$) & \multicolumn{1}{c}{0.1}\\
    Masking Ratio & \multicolumn{1}{c}{0.75}\\
    \midrule
    Random Resized Crop Scale, Ratio & (0.08, 1.0), (0.75, 1.3333) \\
    Interpolation & bicubic \\
    Random Horizontal Flip Probability & 0.5\\
    Rand Augment & n = 2\\
    Random Erasing Probability, Mode and Count & 0.25, Pixel, (1, 1)\\
    Color Jittering & \multicolumn{1}{c}{None}\\
    Auto-augmentation & \multicolumn{1}{c}{rand-m9-mstd0.5-inc1} \\
    Mixup  & \multicolumn{1}{c}{True}\\
    Cutmix  & \multicolumn{1}{c}{False}\\
    Mixup, Cutmix Probability & \multicolumn{1}{c}{1, 0}\\
    Mixup Switch Probability & \multicolumn{1}{c}{0.5}\\
    Mixup Mode  & \multicolumn{1}{c}{Batch}\\
    Label Smoothing  & \multicolumn{1}{c}{0.1}\\
    \bottomrule
    \end{tabular}}
    \label{training_config}
\end{table*}

\section{Training Configurations} \label{training_appendix}
We follow the configurations introduced in MAE~\cite{mae}. A comprehensive set of training configurations for all datasets used in this study is provided in Table~\ref{training_config} for reference. During training two parameters, \textbf{image and patch sizes} vary depending upon the datasets and the rest of the parameters are the same across all the datasets.

\textbf{Swin and ConvMAE training configurations: }
We have adopted the training pipeline from~\cite{uniform_masking} and~\cite{convmae} for Swin~\cite{swinver1} and ConvMAE~\cite{convmae} respectively. For each of them, we have combined their reconstruction based self-supervised learning (SSL) and fine-tuning in a joint learning framework, keeping the training configurations same. Note that UM-MAE~\cite{uniform_masking} with its secondary masking strategy, is an efficient version of SimMIM~\cite{simmim} allowing the reconstruction based SSL in hierarchical transformers like Swin~\cite{swinver1} and PVT~\cite{wang2021pvtv2}.

\section{ViT Decoder for Reconstruction based SSL} \label{decoder_appendix}
In contrast to MAE~\cite{mae}, this paper employs a reconstruction-based SSL approach with class-wise supervision. Consequently, we explore the effect of different design choices of the ViT decoder, which can impact the SSL training while simultaneously optimizing cross-entropy in Self-supervised Auxiliary Task (SSAT). To this end, we conduct experiments that involve modifying three decoder attributes: \textbf{depth}, \textbf{dimension}, and \textbf{attention heads}. We evaluate the resulting impact on the model's top-1 accuracy using two datasets: CIFAR-10 and CIFAR-100.

\textbf{Decoder Depth: }
In this study, we investigated the impact of decoder depth on model performance, as shown in Table~\ref{ablation:decoder_depth}. During the experiments, we maintained a fixed decoder dimension of 128, decoder heads of 16, and a value of $\lambda$ equal to 0.1. Our findings demonstrate that the optimal results for both datasets were obtained at a decoder depth of 2.


\textbf{Decoder Embedding Dimension: }
This section investigates the influence of the decoder embedding dimensions on model performance, as presented in Table~\ref{ablation:decoder_dimension}. Throughout these experiments, we maintained a constant value of $\lambda$ at 0.1, a decoder depth of 2, and 16 decoder heads. Our results indicate that the optimal performance was achieved with a decoder dimension of 128.


\textbf{Decoder Heads: }
Table~\ref{ablation:decoder_heads} presents the outcomes of the ablation study performed to evaluate the impact of the number of heads on the ViT's performance. The hyperparameters, namely $\lambda = 0.1$, decoder$\_$depth $= 2$, and decoder$\_$dimension $= 128$, are fixed to their optimal values from the prior experiments. The experimental findings indicate that retaining 4 heads for CIFAR-10 and 8 heads for CIFAR-100 resulted in the highest performance levels. To ensure generalizability across our experiments, we fixed the number of decoder heads to 16.

\begin{figure*}
\begin{center} 
\includegraphics[scale=0.2]{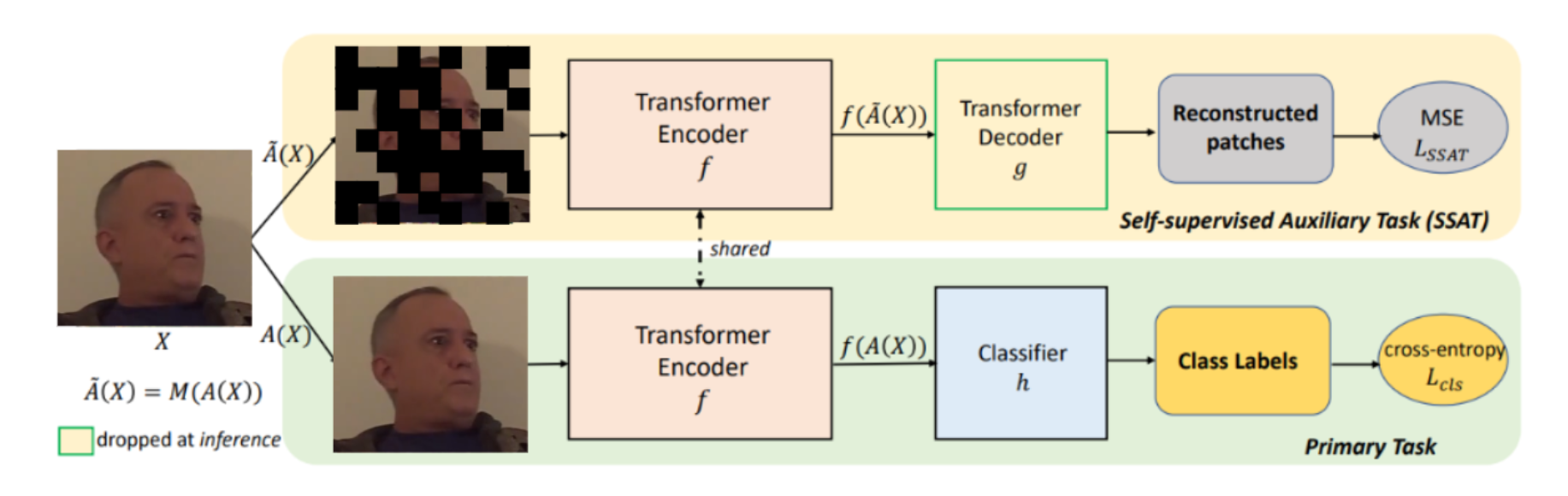}
\caption{Mask Autoencoder as a Self Supervised Auxiliary Task for deepfake detection.}
\label{deepfake_arch}
\end{center} 
\end{figure*}

\section{Details of deepfake detection experiments} \label{deepfake_appendix}
In this section, we elaborate the cross-training manipulation and zero-shot transfer experimental details for deepfake detection.

\subsection{Datasets}
We employ two publicly available popular dataset on Deepfakes.

\textbf{FaceForensics++}: The FaceForensics++ dataset \cite{roessler2019faceforensicspp} is a large-scale benchmark dataset for face manipulation detection, which is created to help develop automated tools that can detect deepfakes and other forms of facial manipulation. The dataset consists of more than 1,000 high-quality videos with a total of over 500,000 frames, which were generated using various manipulation techniques such as facial reenactment, face swapping, and deepfake generation.

The videos in the dataset are divided into four categories, each corresponding to a different manipulation technique: Deepfakes, Face2Face, FaceSwap, and NeuralTextures. Deepfakes use machine learning algorithms to generate realistic-looking fake videos, while Face2Face and FaceSwap involve manipulating the facial expressions and identity of a person in a video. NeuralTextures uses a different approach by altering the texture of a face to make it appear different.
The dataset includes both real and manipulated videos, with each manipulation technique applied to multiple individuals. The statistics of different manipulating techniques available in faceforensics++ is provided in Table~\ref{tab:stat_ff++}.

\textbf{DFDC}: The Deepfake Detection Challenge (DFDC) dataset \cite{seferbekov2020dfdc} is a large-scale benchmark dataset for deepfake detection. The dataset consists of more than 100,000 videos generated using various facial modification algorithms.
The DFDC dataset consists of two versions: a preview dataset with 5k videos featuring two facial modification algorithms and a full dataset with 124k videos featuring eight facial modification algorithms.
The DFDC dataset is the largest currently and publicly available face swap video dataset, with around 120,000 total clips sourced from 3,426 paid actors. The videos are produced using several Deepfake, GAN-based, and non-learning methods. The official DFDC train, validation and test splits are also designed to simulate real-world performance, with the validation set consisting of a manipulation technique not present in the train set, and the test set containing much more challenging augmentations and perturbations.

\subsection{Methodology}
\textbf{VideoMAE} \cite{videomae}: VideoMAE is a self-supervised video pre-training method that extends masked autoencoders (MAE) to videos. VideoMAE performs the task of masked video modelling for video pre-training. It employs an extremely high masking ratio (90\%-95\%) and tube masking strategy to create a challenging task for self-supervised video pre-training.
The temporally redundant video content enables a higher masking ratio than that of images. This is partially ascribed to the challenging task of video reconstruction to enforce high-level structure learning.

\textbf{SSAT}: In this experiment, we use the same backbone as in the original work~\cite{videomae} and we use rigorous augmentations as used by the winners of the DFDC Challenge \cite{seferbekov2020dfdc} in our experimental setting.
For training VideoMAE along with SSAT on DFDC, we extend our image based framework to videos (as illustrated in Figure~\ref{deepfake_arch}) and jointly optimize the primary deepfake classification loss $L_{cls}$ and the auxiliary video reconstruction loss $L_{SSAT}$ as
\begin{equation} \label{eq:ssat_loss}
    L = \lambda * L_{cls} + (1 - \lambda) * L_{SSAT}
\end{equation}
where $\lambda=0.1$ is the loss scaling factor. 

\subsection{Implementation details}
While training VideoMAE+SSAT models follow the training recipe of~\cite{videomae}, we have incorporated specific modifications tailored for deepfake detection.

\textbf{Fake class weight:} Assigns weight $w$ to the class representing \textit{fake} in the weighted cross entropy loss. This was used since the training set is very imbalanced (82\% fake - 18\% real).

\begin{equation} \label{eq:weighted_ce_loss}
    L_{CE} = - ( w t_{real} \log p_{real} + (1-w) t_{fake} \log p_{fake})
\end{equation}
Equation \ref{eq:weighted_ce_loss}: Weighted Cross-Entropy Loss. $w$ is the weight of the \textit{real} class while $p_{real}$ and $p_{fake}$ are the predicted probabilities, and $t_{real}$ and $t_{fake}$ are the ground truth indicator variables.

\textbf{Augmentations:} The choice of augmentations has a profound impact on validation performance. The set of augmentations that work best are Image Compression, Gaussian Noise, Gaussian Blur, Horizontal Flip, Brightness Contrast, FancyPCA, Hue Saturation, Greyscale and shift-scale-rotate, all available in the Albumentations library \cite{buslaev2020albumentations} and used in the DFDC challenge’s winning solution by Selim Seferbekov \cite{seferbekov2020dfdc}. Other augmentations like Reversal, Random up / down sampling and heavy Gaussian Noise seem to have a detrimental effect, possibly because they do not generalize to the validation set. Meanwhile, having no augmentations also decreases the generalizability.
 
\textbf{Testing:} During testing, predictions are obtained by averaging the results from all 16-frame segments across the entire video.

\section{Attention Visualization} \label{vis_appendix}
\begin{figure*}[!ht]
     \centering
     \begin{subfigure}[b]{0.21\textwidth}
         \centering
         \includegraphics[width=0.9\textwidth]{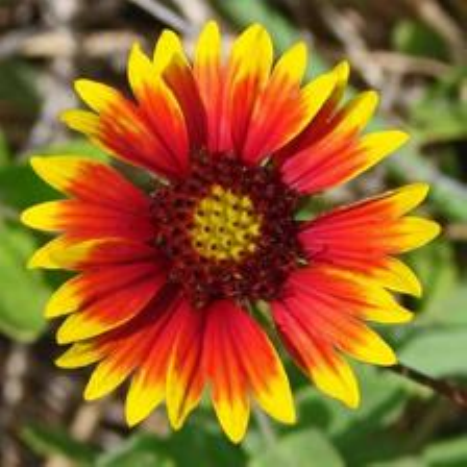}
     \end{subfigure}
     \hfill
     \begin{subfigure}[b]{0.21\textwidth}
         \centering
         \includegraphics[width=0.9\textwidth]{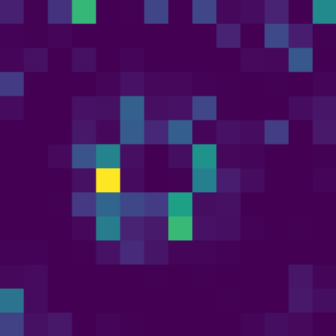}
     \end{subfigure}
     \hfill
     \begin{subfigure}[b]{0.21\textwidth}
         \centering
         \includegraphics[width=0.9\textwidth]{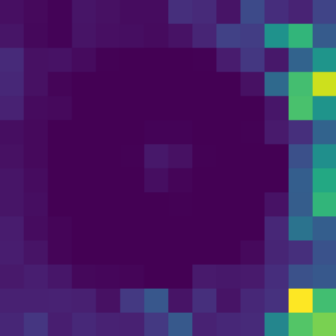}
     \end{subfigure}
     \hfill
     \begin{subfigure}[b]{0.21\textwidth}
         \centering
         \includegraphics[width=0.9\textwidth]{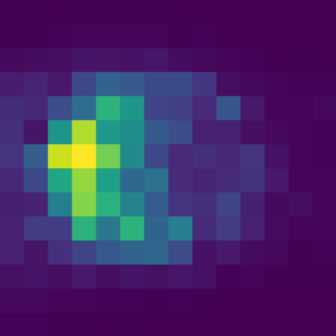}
     \end{subfigure}
    
     \begin{subfigure}[b]{0.21\textwidth}
         \centering
         \includegraphics[width=0.9\textwidth]{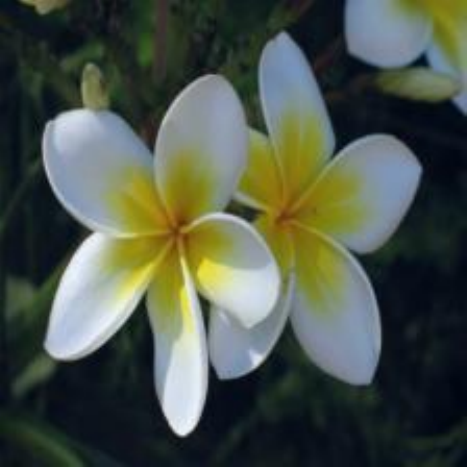}
     \end{subfigure}
     \hfill
     \begin{subfigure}[b]{0.21\textwidth}
         \centering
         \includegraphics[width=0.9\textwidth]{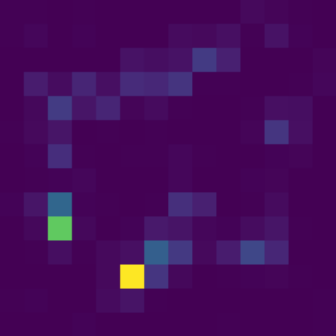}
     \end{subfigure}
     \hfill
     \begin{subfigure}[b]{0.21\textwidth}
         \centering
         \includegraphics[width=0.9\textwidth]{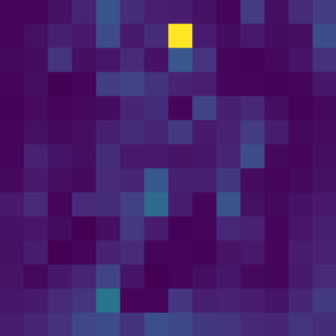}
     \end{subfigure}
     \hfill
     \begin{subfigure}[b]{0.21\textwidth}
         \centering
         \includegraphics[width=0.9\textwidth]{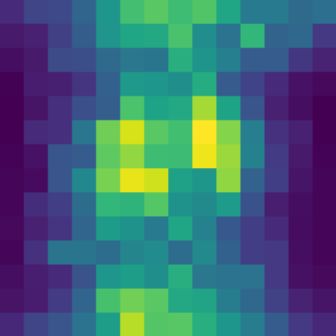}
     \end{subfigure}

     \begin{subfigure}[b]{0.21\textwidth}
         \centering
         \includegraphics[width=0.9\textwidth]{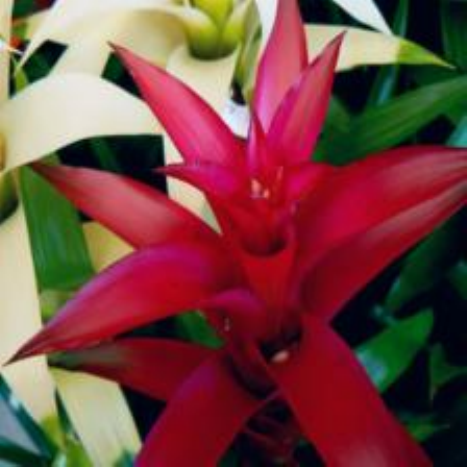}
     \end{subfigure}
     \hfill
     \begin{subfigure}[b]{0.21\textwidth}
         \centering
         \includegraphics[width=0.9\textwidth]{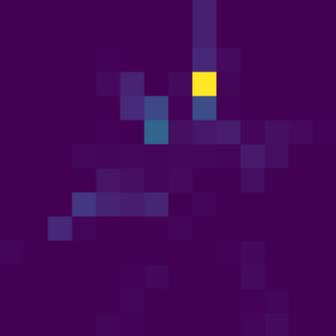}
     \end{subfigure}
     \hfill
     \begin{subfigure}[b]{0.21\textwidth}
         \centering
         \includegraphics[width=0.9\textwidth]{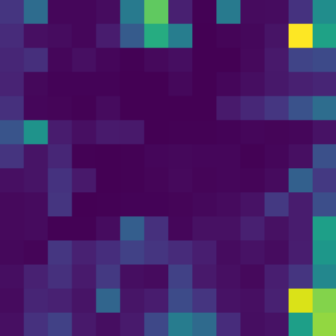}
     \end{subfigure}
     \hfill
     \begin{subfigure}[b]{0.21\textwidth}
         \centering
         \includegraphics[width=0.9\textwidth]{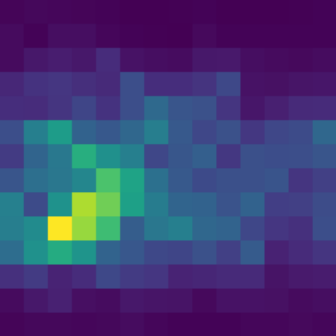}
     \end{subfigure}

     \begin{subfigure}[b]{0.21\textwidth}
         \centering
         \includegraphics[width=0.9\textwidth]{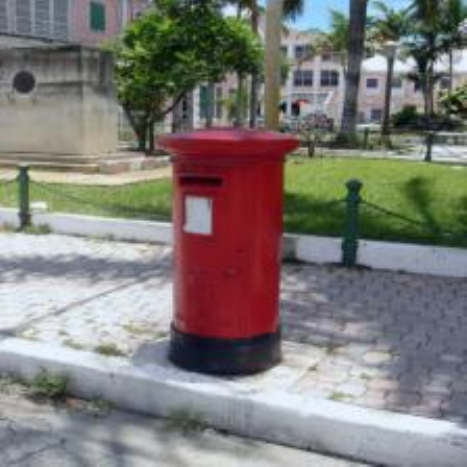}
     \end{subfigure}
     \hfill
     \begin{subfigure}[b]{0.21\textwidth}
         \centering
         \includegraphics[width=0.9\textwidth]{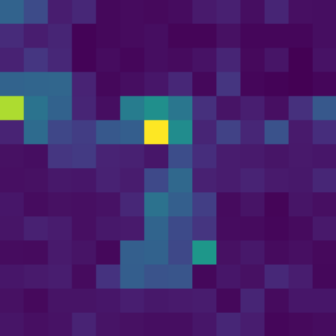}
     \end{subfigure}
     \hfill
     \begin{subfigure}[b]{0.21\textwidth}
         \centering
         \includegraphics[width=0.9\textwidth]{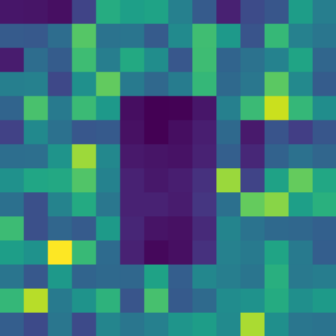}
     \end{subfigure}
     \hfill
     \begin{subfigure}[b]{0.21\textwidth}
         \centering
         \includegraphics[width=0.9\textwidth]{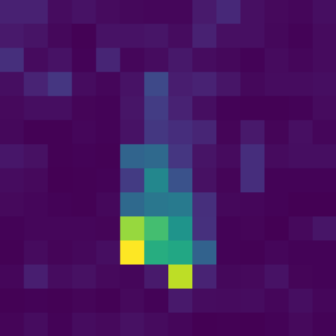}
     \end{subfigure}

     \begin{subfigure}[b]{0.21\textwidth}
         \centering
         \includegraphics[width=0.9\textwidth]{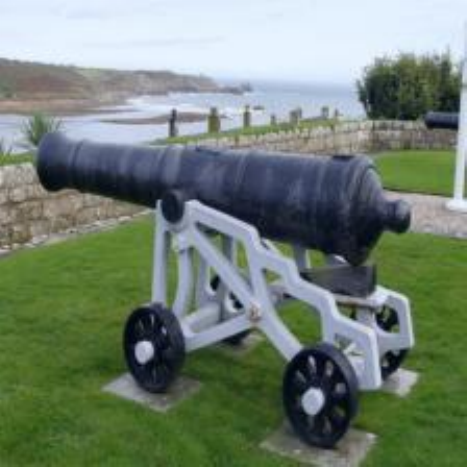}
     \end{subfigure}
     \hfill
     \begin{subfigure}[b]{0.21\textwidth}
         \centering
         \includegraphics[width=0.9\textwidth]{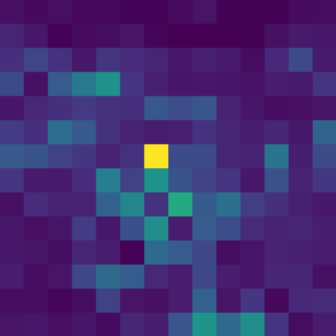}
     \end{subfigure}
     \hfill
     \begin{subfigure}[b]{0.21\textwidth}
         \centering
         \includegraphics[width=0.9\textwidth]{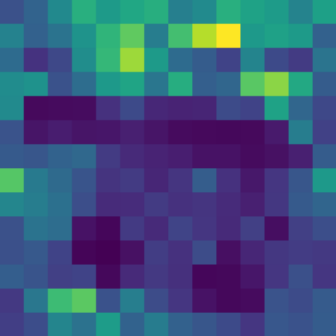}
     \end{subfigure}
     \hfill
     \begin{subfigure}[b]{0.21\textwidth}
         \centering
         \includegraphics[width=0.9\textwidth]{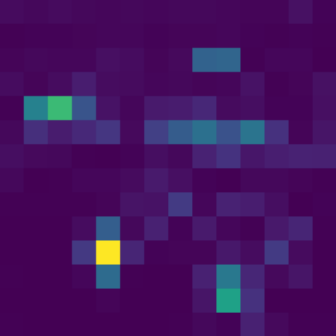}
     \end{subfigure}
    
    \begin{subfigure}[b]{0.21\textwidth}
         \centering
         \includegraphics[width=0.9\textwidth]{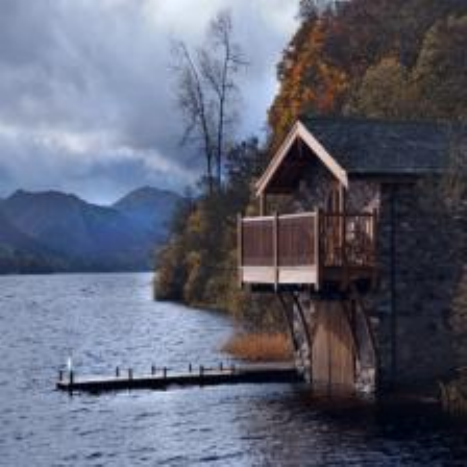}
     \end{subfigure}
     \hfill
     \begin{subfigure}[b]{0.21\textwidth}
         \centering
         \includegraphics[width=0.9\textwidth]{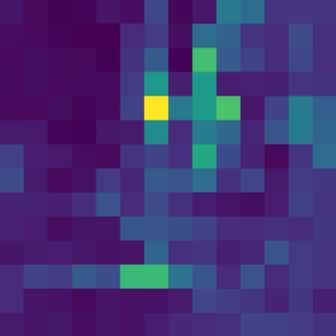}
     \end{subfigure}
     \hfill
     \begin{subfigure}[b]{0.21\textwidth}
         \centering
         \includegraphics[width=0.9\textwidth]{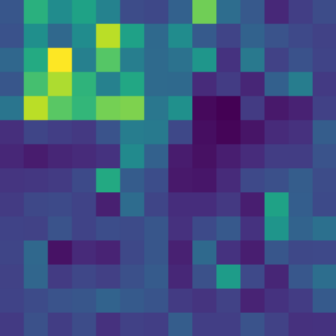}
     \end{subfigure}
     \hfill
     \begin{subfigure}[b]{0.21\textwidth}
         \centering
         \includegraphics[width=0.9\textwidth]{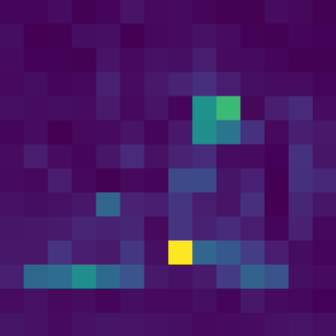}
     \end{subfigure}
    \caption{Attention visualization of six images, three from the Oxford Flowers-102 dataset (top 3 rows) and three from the ImageNet dataset (bottom 3 rows). The attention heatmaps in the second, third, and fourth columns correspond to models trained from scratch using ViT, models trained using SSL+FT, and models trained using SSAT, respectively.}
    \label{attention}
\end{figure*}

In Figure~\ref{attention}, we illustrate attention visualization for a few sample images drawn from the Flower and ImageNet datasets. Our analysis of the visualization highlights that the ViT trained with SSAT generates attention maps that emphasize the primary object class to a greater extent than the attention maps computed by ViTs trained from scratch and trained with SSL+FT. These findings indicate that the ViT trained with SSAT exhibits higher efficacy in image classification.

\end{document}